\documentclass{article} 
 \usepackage[preprint]{neurips_2023}
\usepackage[table, svgnames, dvipsnames]{xcolor}

\usepackage{amsmath,amsfonts,bm}

\def\eqref#1{equation~\ref{#1}}

\def\floor#1{\lfloor #1 \rfloor}
\def\1{\bm{1}}

\DeclareMathAlphabet{\mathsfit}{\encodingdefault}{\sfdefault}{m}{sl}
\SetMathAlphabet{\mathsfit}{bold}{\encodingdefault}{\sfdefault}{bx}{n}

\usepackage{longtable}
\usepackage{tabularx}
\usepackage{tcolorbox}
\usepackage{hyperref}
\usepackage{url}
\usepackage{tablefootnote}
\usepackage{booktabs}
\usepackage{graphicx}
\usepackage{subcaption}

\title{Learning from Contrastive Prompts: Automated Optimization and Adaptation}

\author{Mingqi Li~\thanks{This work was done during author's internship at Amazon.} \\
Clemson University\\
\texttt{mingqil@clemson.edu}
        \And 
        Karan Aggarwal \\ Amazon Services LLC \\ \texttt{kagg@amazon.com}
        \And
        Yong Xie \\ Amazon Services LLC \\ \texttt{yonxie@amazon.com}
        \And
        Aitzaz Ahmad \\ Amazon Services LLC \\ \texttt{aitzaza@amazon.com}
        \And
        Stephan Lau \\ Amazon Services LLC \\ \texttt{lausteph@amazon.com}
}

\begin{document}

\maketitle

\begin{abstract}


As LLMs evolve, significant effort is spent on manually crafting prompts. While existing prompt optimization methods automate this process, they rely solely on learning from incorrect samples, leading to a sub-optimal performance. Additionally, an unexplored challenge in the literature is prompts effective for prior models may not perform well on newer versions or different languages. We propose the Learning from Contrastive Prompts (LCP) framework to address these gaps, enhancing both prompt optimization and adaptation. LCP employs contrastive learning to generate effective prompts by analyzing patterns in good and bad prompt examples. Our evaluation on the Big-Bench Hard dataset shows that LCP has a win rate of over 76\% over existing methods in prompt optimization and demonstrates strong adaptability across different model versions, families, and languages. LCP offers a systematic approach to prompt engineering, reducing manual effort in deploying LLMs across varied contexts.

\end{abstract}

\section{Introduction}

The current approach to utilize Large Language Models (LLMs) begins with users providing their queries. These queries are then augmented with additional instructions, called prompts, by the system to enhance response quality. Prompts often include contextual information or instructions that help the model better understand and respond to a query. Prompts may also include guidelines to restrict the LLM from generating harmful or inappropriate content, ensuring safer and more reliable interactions. The process of writing these prompts typically involves trial-and-error. This intermediate step, known as \emph{prompt engineering}, is crucial for optimizing the performance of LLMs.

Recent advancements in prompt engineering have introduced various techniques to enhance the effectiveness of prompts. One notable example is zero-shot chain-of-thought prompting~\citep{kojima2022large}, where simply adding the phrase ``Let's think step-by-step" can stimulate the LLMs' reasoning capabilities, encouraging it to think aloud and process the query in a logical sequence. However, this seemingly magical and straightforward phrase is hard to come up with, as current LLMs are very sensitive to phrasing prompts. Semantically similar prompts can lead to significant performance variations (\citealp{kojima2022large}; \citealp{zhou2023large}; \citealp{salinas2024butterfly}), 
with minor modifications resulting in a performance drop. This variation requires numerous experiments to find the optimal prompt, resulting in a labor-intensive and time-consuming prompt engineering process.

Prior works in literature (\citealp{yang2024large}; \citealp{guo2024connecting}; \citealp{wang2024promptagent}; \citealp{zhou2023large}; \citealp{sun2023autohint}) have addressed these limitations by automatically optimizing prompts. For instance, AutoHint~\citep{sun2023autohint} proposed learning from wrong samples by using LLMs to generate hints for selected incorrect samples, which are used to refine prompts. 
However, learning from only incorrect samples can make the prompts too specific to the wrong samples, losing an understanding of what worked.
Another approach by OPRO~\citep{yang2024large} involves using LLMs as optimizers, where the model generates new prompts iteratively based on a ranking list of previous prompts and their corresponding scores. However, OPRO lacks the incorporation of feedback from incorrect samples, potentially limiting its ability to achieve optimal performance.

While prompt optimization has prior art, an unexplored and significant challenge in prompt engineering is \emph{prompt adaptation}. As LLMs are continually updated and more capable LLMs are introduced, existing prompts often need to be rewritten and tailored to align with the new model version or an entirely new model. This constant adaptation is necessary to maintain the effectiveness of the prompts to ensure they produce high-quality results. Additionally, prompt adaptation across various languages is crucial for ensuring performance in multilingual contexts. However, this area remains unexplored in the literature.




To address these gaps, we propose \emph{Learning from Contrastive Prompts (LCP)}, an automatic prompt optimization and adaptation framework. In particular, our framework consists of two stages: prompt candidate generation and new prompt generation. We inject diverse prompts into prompt optimization by generating multiple prompt candidates to explore the prompt space. To overcome the shortcomings of existing methods, we take an inspiration from the principle of contrastive learning~\citep{chen2020simple} by allowing the LLM to contrast between good and bad prompts from the generated prompt candidates while learning to improve on error cases. This helps the LLM reason on the prompts that work versus those that do not, using exploration, to incorporate good prompts without being too specific to the error cases.

We demonstrate the effectiveness of our approach for both the scenarios of prompt optimization and prompt adaptation. We evaluate our framework on the Big-Bench Hard dataset~\citep{suzgun2022challenging}, which comprises diverse tasks considered challenging even for human evaluators. Our framework achieves a win rate of over 76\% versus OPRO~\citep{yang2024large} and AutoHint~\citep{sun2023autohint} on prompt optimization. It especially excels at algorithmic and multi-step arithmetic reasoning tasks. 
Our prompt adaptation framework leverages feedback from the target model to enhance performance on the prompts from the source model. It achieves comparable or better results than prompt optimization from scratch on the target model when the target model is a weaker model. Our results show that prompt adaptation is a delicate balance between the target model's abilities and the source model's abilities. It can slightly degrade performance on tasks where the source model excels, while improving performance on tasks where the target model is stronger. This observation holds true across model versions and families, \emph{with our framework creating a balance between the strengths of the source and target models}. Results on the XCOPA dataset further demonstrate our framework's capability to adapt prompts across languages with a better performance on 7 out of 11 languages versus prompt refinement baselines, especially for low resource languages like Swahilli and Southern Quechua. 

In summary, we present a novel framework using contrastive learning for prompt optimization and an unexplored problem of prompt adaptation. Our results show promising results on both the prompt optimization and prompt adaptation across model versions, families, and languages.

\section{Methodology}

\begin{figure}[h]
  \centering
    \includegraphics[width=12cm]{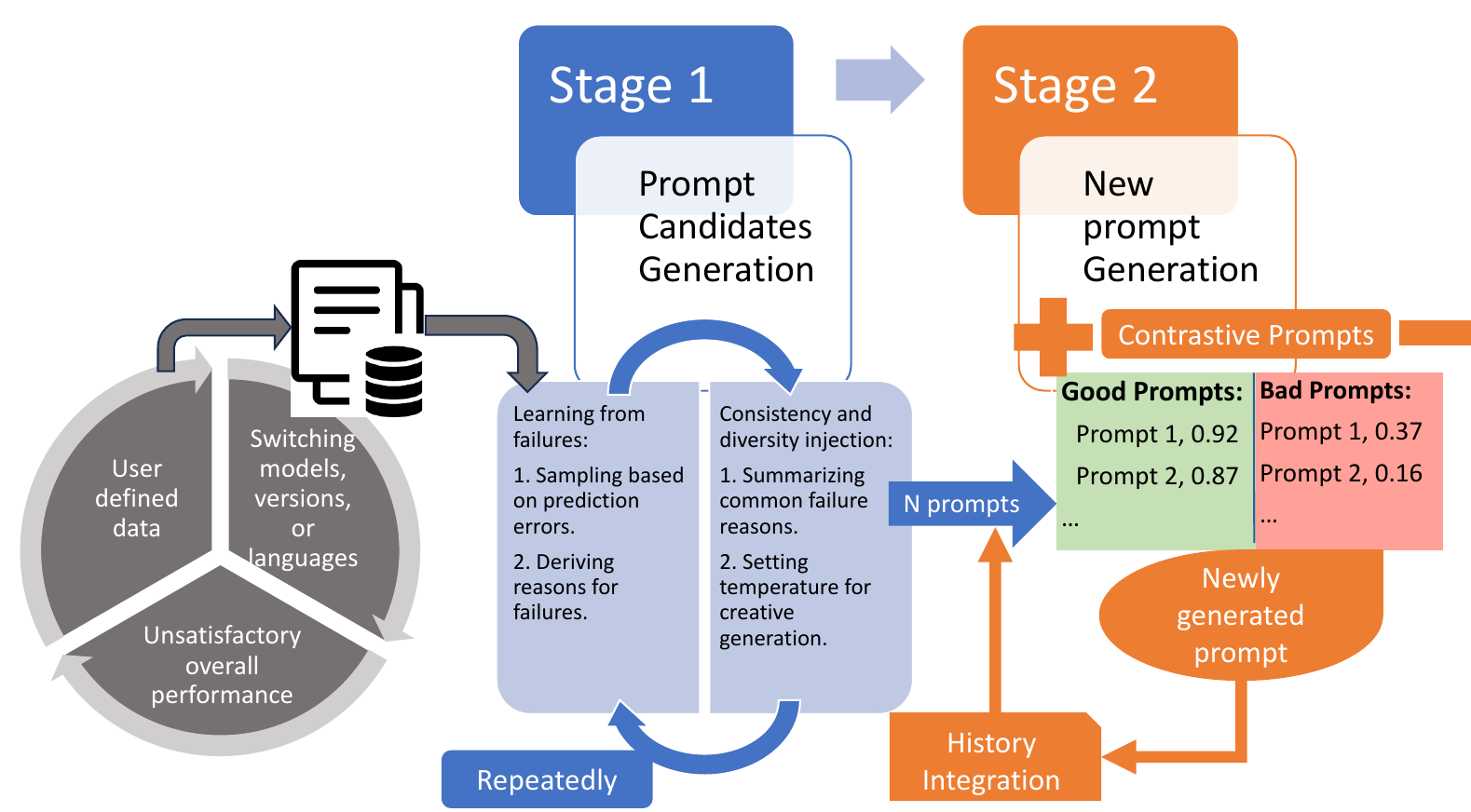}
  \caption{Learning from Contrastive Prompts (LCP) framework. Given an initial prompt and a small training set, LCP generates multiple prompt candidates derived from summaries of common failure reasons for different sample combinations. It leverages the inherent capabilities of LLMs to understand the underlying patterns through contrastive prompts to generate a new prompt.}
  \label{figure:framework}
  \vspace{-1.5em}
\end{figure}

Our proposed framework is illustrated in Figure~\ref{figure:framework}. Both prompt optimization and prompt adaptation utilize the same framework with minor modifications. The framework is designed to enhance the effectiveness of prompts across various scenarios, including adapting to different model versions, model families, and languages. In this section, we will explain the processes of prompt optimization and prompt adaptation separately, detailing the specific stages and mechanisms involved in each.


\subsection{Motivation}

In line with recent advancements, our work harnesses the reasoning capabilities of LLMs to automatically optimize prompts. However, a significant challenge persists in effectively instructing LLMs to maximize their potential for generating high-quality prompts. 

Our approach emulates how humans learn: by understanding failures and their reasons, as well as contrasting good and bad examples to grasp what works and what does not. We provide LLMs with error case feedback and a spectrum of prompt quality. We expose them to failures, their reasons, and ask them to contrast between good and bad prompts. This enables learning from diverse perspectives for an improved prompt generation. This unexplored avenue holds significant potential for gaining a comprehensive understanding and unlocking LLMs' full capabilities in generating high-quality prompts.

The concept of contrasting high-quality prompts with low-quality ones draws parallels to the principles of contrastive learning, which has gained significant traction across various domains~\citep{chen2020simple}. Contrastive learning aims to learn meaningful representations by maximizing the similarity between positive pairs (analogous to high-quality prompts) while minimizing the similarity between negative pairs (analogous to low-quality prompts). By leveraging contrastive learning techniques in our context, we can potentially guide LLMs to capture the essential characteristics of effective prompts. In particular, given a list of prompts with their corresponding quality scores, we compare a batch of high-quality prompts to a batch of low-quality prompts, drawing conclusions about the patterns that characterize effective prompts.

\subsection{Prompt Optimization}

Prompt optimization improves the performance of prompts starting with a simple initial prompt and iteratively refining it to enhance the performance of the LLM on the tasks at hand. We present our two main components next: Prompt Candidate Generation to generate candidate prompts and New Prompt Generation to generate a final prompt using the insights from the candidate prompts.  

\subsubsection{Prompt Candidate Generation}

 We start with a small training set and an initial prompt, which is appended to each query. Given an input and its expected output, LLMs are capable of understanding how to achieve the expected output. We evaluate the LLM generated outputs on the small training set to identify failure cases. Taking a motivation from AutoHint~\citep{sun2023autohint}, we use LLMs to learn from failures by understanding the failures, and generating reasons behind the failures. The prompt template in this step is shown in Appendix~\ref{sec:meta_prompts}.

\textbf{Consistency and diversity injection.} A problem with generating the reason for each wrong sample and crafting a prompt candidate based on it, is that the candidate prompt can be biased towards that sample, making it too specific. To inject some consistency, we select multiple incorrectly predicted samples and summarize the common failure reasons (See Appendix~\ref{sec:meta_prompts}). \emph{AutoHint also summarized feedback on incorrect samples, but they directly used these summaries as new prompts, which led to their performance being significantly influenced by the selected samples.} This presents a dilemma; if we select samples which are quite similar, this could lead to model over-fitting on these samples,  trapping the generated prompt in a local minimum. To address this, we add diversity to the generated summary by setting a more creative temperature and repeating this step multiple times to generate $N$ prompts, referred to as \emph{prompt candidates}. This approach helps the model to explore the prompt space. We use $N=10$ based on our experiments.

\textbf{History integration.} The generated prompts from previous iterations can also influence the optimization process, leading to a better performance. Therefore, we integrate these prompts into the prompt candidates, ensuring that the accumulated knowledge from past iterations contributes to the ongoing optimization process.

\subsubsection{New Prompt Generation}

Now that we have $N$ prompt candidates, our goal is to generate a new prompt using them. First, we assign a score to each candidate based on its inference performance on the training set. We then rank all the candidates according to their scores. Inspired by contrastive learning, we instruct the LLM to identify the underlying patterns that distinguish good prompts from bad prompts. Specifically, we define the top-$K$ prompts as the good prompts and the bottom-$K$ prompts as the bad prompts, and we use the meta-prompt shown in Appendix~\ref{sec:meta_prompts} to instruct the LLM to generate a new prompt that follows the underlying pattern of good prompts while improving the performance. 

\emph{This approach simplifies the learning process for LLMs compared to OPRO, which directly learns from a ranked list and also did not provide any error case feedback.} Viewing batches of good or bad prompts can also be considered as injecting the consistency. Additionally, since we integrate prompts generated in previous iterations, the differentiation between good and bad prompts becomes more pronounced over time.



\subsection{Prompt Adaptation}

Prompt adaptation addresses the need to switch between different model versions, model families, or languages, ensuring that optimized prompts remain effective across diverse conditions. While our main framework remains unchanged, we will detail the specific strategies used for adaptation.

\begin{figure}[h]
  \centering
    \includegraphics[width=10cm]{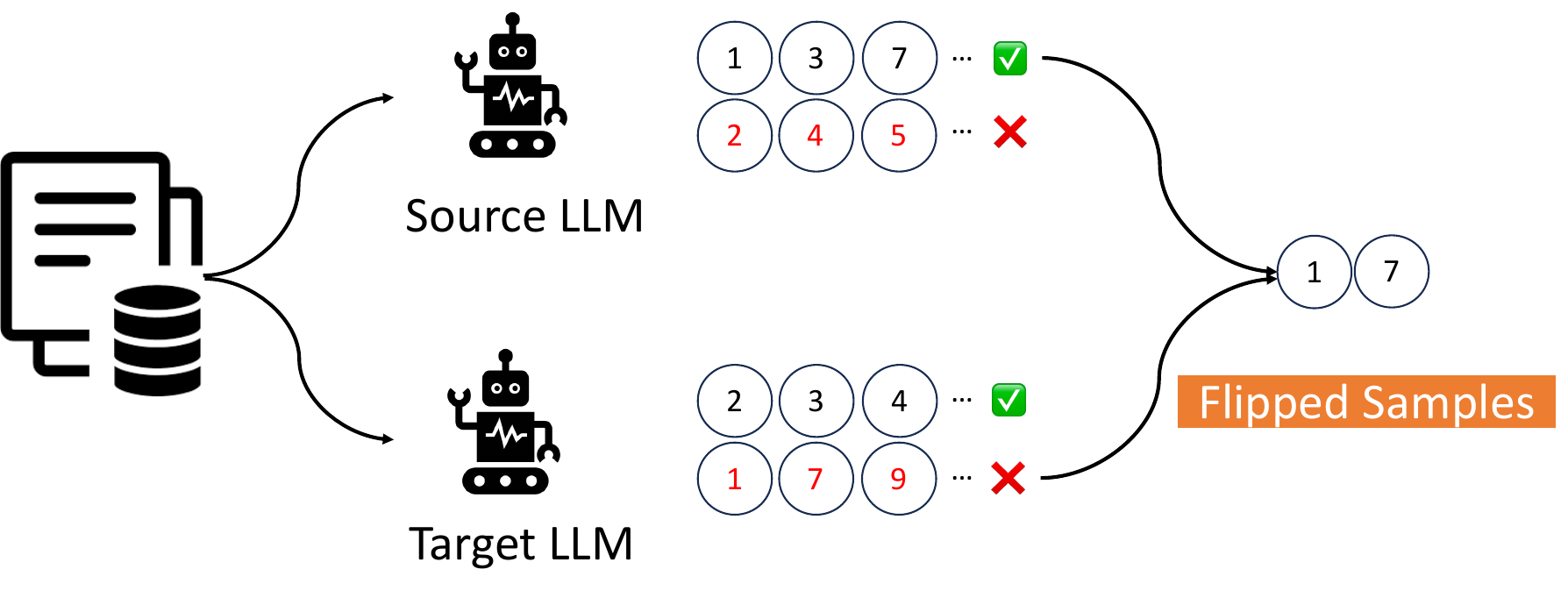}
  \caption{Data sampling strategy for prompt adaptation: Run inference on the training set using the source and target LLMs. Identify correctly and incorrectly predicted samples for each model. Select samples correctly predicted by the prior model but incorrectly predicted by the current model.}
  \label{figure:flipping}
\end{figure}

\textbf{Cross-model version.} In real-world scenarios, underlying models are being continuously enhanced with new version roll-outs every few months. For example, the GPT family has evolved to include ChatGPT-4o, and Meta recently released LLAMA-3. Users can address their issues by switching to a more advanced model when the previous version lacks the required capabilities. Conversely, they can revert to the less capable model version considering costs.

Our goal is to refine an optimized prompt for one model version to adapt it effectively to another model version. Integrating feedback from the original model version can provide valuable insights, enhancing the adaptation process for the new model version. To facilitate this, we improve our data sampling strategy (see Figure~\ref{figure:flipping}). \textbf{Instead of randomly selecting incorrectly predicted samples, we focus on samples that are correctly predicted by the original model version but incorrectly predicted by the current model version}. This ensures the LLM pays more attention to these critical samples, leading to a more effective adaptation.

\textbf{Cross-model.} Next, we want to take it a step further. Users may want to switch to models from other families that are more accessible or have proven to be more effective for their specific tasks. This setting is more challenging because the underlying models are fine-tuned with different data distributions, tasks, and instructions, resulting in significant variations. In this scenario, we maintain the same data sampling strategy used in the cross-model version, while incorporating error tolerance through wrong format rejection to accommodate less effective models like LLAMA. In particular, if the generated prompt does not adhere to the defined output format, we regenerate it until we reach the maximum allowable number.

\textbf{Cross-lingual.} Adapting prompts across different languages presents unique difficulties due to variations in linguistic structures, vocabularies, and resources. To simplify the process and provide a universal approach, we extend the same strategy to handle prompt adaptation across languages, demonstrating its broader applicability. In particular, we have the LLM translate samples from the target languages into English and then conduct the inference step. We select data samples that are correctly predicted when translated into English but incorrectly predicted in their original languages.

\section{Experiments}
\vspace{-1em}

\subsection{Setup}

\textbf{Benchmarks.} Our evaluation benchmark is a subset of the Big-Bench Hard dataset~\citep{suzgun2022challenging}, consisting of 17 challenging multi-choice tasks. The tasks are diverse, spanning across various categories like natural language understanding, the use of world knowledge (both general and factual), multilingual knowledge and reasoning, and algorithmic and multi-step arithmetic reasoning, making it a comprehensive test for our framework. We report results for each  task category based on the keyword taxonomy provided by Big-Bench Hard dataset\footnote{\url{https://github.com/google/BIG-bench/blob/main/keywords.md}}. For the cross-lingual setting, we use the XCOPA dataset~\citep{ponti-etal-2020-xcopa}, which demonstrates common sense reasoning ability and requires world knowledge understanding. 

\textbf{Models.} For our experiments, we used several state-of-the-art LLMs to evaluate the effectiveness of our framework. These models include: Claude-3-sonnet~\citep{anthropic2024claude} and Claude-3-haiku, LLAMA-3-70b~\citep{dubey2024llama}. Claude-3-haiku is a smaller and faster model while Claude-3-sonnet is a more powerful model on the leaderboards\footnote{\url{https://chat.lmsys.org/?leaderboard}}. 

\textbf{Baselines.} We compare our framework with two existing methods: AutoHint~\citep{sun2023autohint} and OPRO~\citep{yang2024large}. AutoHint optimizes prompts based on wrong samples in two iterations, using hint generation and summarization. OPRO optimizes prompts by maintaining a ranking list of historical prompts and relying solely on that. Since these works used LLMs such as the GPT family and the PaLM family, which we don't have access to, we reimplemented their techniques on our target LLMs for a fair comparison.

\textbf{Prompt selection strategy.} Our framework and OPRO both involve optimizing prompts iteratively, which can lead to performance fluctuations even upon convergence. Additionally, each task from the Big-Bench Hard dataset consists of only 250 samples, making it infeasible to create a validation set. This limitation is consistent with real-world scenarios where data availability is often restricted. We simply use the prompt generated in the last iteration and also present the performance of the prompt with the best training set accuracy during the process.

\textbf{Implementation details} We use the same data split as OPRO on the Big-Bench Hard dataset, with 50 samples for training and 200 samples for testing. For the XCOPA dataset, we use 50 samples from the validation set for training, and test on 500 samples from the original testing set. The temperature is set to 1.0. The maximum number of iterations is set to 50, followed by a selection step. In each random sampling step, we select 3 incorrectly predicted samples and repeat this step 10 times. For contrastive prompts, we select 3 good prompts and 3 bad prompts from the ranking list.

\begin{table}[t!]
\small
\caption{Test accuracy of prompt optimization approaches on four types of 17 BBH tasks for last iteration prompt (Last) versus the prompt with best training accuracy (Best). Reported results are average across 5 runs. - indicates cases where AutoHint could not produce any meaningful results. \textcolor{blue}{Blue} indicates overall best results for Last or Best. \textbf{Bold} indicates highest value in a row. Win rates have been calculated with pair-wise comparisons following ~\cite{liang2022holistic}.}
\label{table:op_perf}
\begin{center}
\begin{tabular}{lccc}
\hline
\multicolumn{1}{c}{\bf TASK}  &\multicolumn{1}{c}{\bf LCP} &\multicolumn{1}{c}{\bf AutoHint} &\multicolumn{1}{c}{\bf OPRO} \\
\hline
& \multicolumn{1}{c}{Last / Best} & \multicolumn{1}{c}{Last / Best} & \multicolumn{1}{c}{Last / Best} \\
\hline
\rowcolor{orange!15}
\emph{Algorithmic and Multi-Step Arithmetic Reasoning} & & & \\
\hline
geometric\_shapes & \textbf{51.25} / \textbf{61.00} & 45.00 / 45.00 & 33.50 / 33.50 \\
logical\_deduction\_three\_objects & \textbf{92.50} / \textbf{90.25}  & 76.00 / 76.00  & 70.50 / 76.00 \\
logical\_deduction\_five\_objects & \textbf{71.50} / \textbf{74.50}  & 52.00 / 52.00  & 54.00 / 65.00 \\
logical\_deduction\_seven\_objects & \textbf{62.00} / \textbf{62.50} & 2.00 / 2.00  & 48.00 / 48.50 \\
penguins\_in\_a\_table & \textbf{97.86} / \textbf{96.15} &  86.30 / 86.30 & 87.20 / 94.00 \\
reasoning\_about\_colored\_objects & 85.30 / 84.40 & 85.50 / 85.50  & \textbf{90.50} / \textbf{90.50} \\
temporal\_sequences &  \textbf{98.25} / \textbf{97.00} & - / - & 77.50 / 80.00 \\
tracking\_shuffled\_objects\_three\_objects & 92.00 / 99.00 & - / - & \textbf{99.50} / \textbf{99.50} \\
tracking\_shuffled\_objects\_five\_objects & \textbf{90.50} / \textbf{95.50} & - / - & 82.50 / 89.50 \\
tracking\_shuffled\_objects\_seven\_objects & \textbf{92.10} / \textbf{98.30} & - / - & 72.50 / 92.00 \\
\hline
\rowcolor{gray!15}
Win Rate (\%) & \textbf{81.00} / \textbf{81.00} & 25.00 / 17.00 & 38.00 / 38.00  \\
\hline
\rowcolor{orange!15}
\emph{Natural Language Understanding} & & & \\
\hline
disambiguation\_qa & \textbf{66.83} / \textbf{66.33}  &  55.00 / 57.00 & 50.00 / 50.00 \\
hyperbaton & \textbf{78.25} / \textbf{84.00}  & 63.00 / 63.00  & 29.00 / 42.50 \\
salient\_translation\_error\_detection\tablefootnote{Note, salient\_translation\_error\_detection task comes under both Natural Language Understanding and Multilingual Knowledge and Reasoning but is only counted once in the overall win rates.} & 57.25 / \textbf{69.50} & \textbf{65.00} / 67.00  & 63.00 / 66.50 \\
snarks & 65.73 / 70.98 & \textbf{84.40} / \textbf{84.40} & 67.80 / 67.80 \\
\hline
\rowcolor{gray!15}
Win Rate (\%) & 50.00 / \textbf{88.00} & \textbf{75.00} / 62.00 & 25.00 / 0.00 \\
\hline
\rowcolor{orange!15}
\emph{Use of World Knowledge} & & & \\
\hline
date\_understanding & 75.50 / 74.50  &  75.50 / 75.50 & \textbf{80.50} / \textbf{80.50} \\
movie\_recommendation & \textbf{87.75} / \textbf{85.50} & 72.00 / 72.00  & 36.00 / 36.00 \\
ruin\_names & 76.50 / 75.25 & \textbf{76.50} / \textbf{79.50}  & 68.00 / 68.00 \\
\hline
\rowcolor{gray!15}
Win Rate (\%) & \textbf{50.00} / 50.00 & 33.00 / \textbf{67.00} & 33.00 / 33.00    \\
\hline
\rowcolor{orange!15}
\emph{Multilingual Knowledge and Reasoning} & & & \\
\hline
salient\_translation\_error\_detection & 57.25 / \textbf{69.50} & \textbf{65.00} / 67.00  & 63.00 / 66.50 \\
\hline
\rowcolor{gray!15}
Win Rate (\%) & 0.00 / \textbf{100.0} & \textbf{100.0} / 50.00 & 50.00 / 0.00 \\
\hline
\rowcolor{green!15}
\textbf{Overall Win Rate (\%)} & \textcolor{blue}{\textbf{66.67}} / \textcolor{blue}{\textbf{76.67}} & 42.31 / 42.31 & 33.33 / 26.67 \\ \hline
\end{tabular}
\end{center}
\end{table}


\subsection{Results}


\subsubsection{Prompt Optimization}

Our prompt optimization begins with the initial prompt ``Let's solve the problem." in the same fashion as OPRO and AutoHint. All the experiments in this section are conducted using Claude-3-sonnet to ensure better performance. We report the results on last iteration from our method and baselines as well as from the prompt with best performance on the training set. Either choice is in line with previous works (~\cite{yang2024large, sun2023autohint}) as strategies like a separate validation set, does not provide any benefit owing to being highly correlated with training performance. Also, we did not observe over-fitting with LCP. For further discussion, please refer to Appendix~\ref{appendix:valid}. While we report results from both, given a relatively high variation and a slightly lower performance using the last prompt (47\% win rate versus 53\% for best training set prompt), we recommend using best prompt on the training set as the selected prompt.

As shown in Table~\ref{table:op_perf}, our LCP framework achieves the best performance with a win rate of 66.67\% compared to AutoHint and OPRO when using the last iteration prompt, and 76.67\% when using the best prompt on training set. We believe that the reason for LCP's superior performance over AutoHint is that our method overcomes the limitation of AutoHint which struggles to summarize diverse hints. In contrast to OPRO, we take advantage of LLMs' inherent capability to contrast good prompts and bad prompts, making the process easier and more detailed than relying on a ranked list. Using contrastive learning directly aligns with the way LLMs are fine-tuned with preference modeling, by learning to rank and distinguish between better and worse options~\citep{rafailov2024direct}. Additionally, we pay more attention to failures than OPRO, which solely relies on the generated prompts and their corresponding scores. The results highlight our framework's strong performance particularly on algorithmic and multi-step arithmetic reasoning tasks. This is understandable as algorithmic and arithmetic tasks involve more detailed instructions versus the other three categories which LCP excels through its contrastive and diversity injection mechanisms. Evidence of this is presented in the ablation study which shows that contrastive and diversity injection mechanisms help especially on algorithmic and arithmetic tasks.  


\begin{table}[t]
\small
\caption{Win rates of prompt adaptation and prompt optimization on model version adaptation and model family adaptation: accuracy on BBH tasks for last iteration prompt (Last) versus the best prompt on the training set (Best). \textcolor{blue}{Blue} indicates overall best results for Last or Best.} 
\label{table:adaptation_summary}
\begin{center}
\resizebox{\textwidth}{!}{
\begin{tabular}{lccc}
\hline
\multicolumn{1}{c}{\bf Source $\rightarrow$ Target}  &\multicolumn{1}{c}{\bf LCP Adaptation} &\multicolumn{1}{c}{\bf LCP Optimization on Target} &\multicolumn{1}{c}{\bf Source Optimized}\\
\hline
& \multicolumn{1}{c}{Last / Best} & \multicolumn{1}{c}{Last / Best} & \multicolumn{1}{c}{Last / Best}\\
\hline
\rowcolor{orange!20}
\textbf{Claude 3 Haiku $\rightarrow$ Claude 3 Sonnet (Table ~\ref{table:cross_version_perf_sonnet})} & & &\\
\hline

Algorithmic and Multi-Step Arithmetic Reasoning & 55.00 / 50.00 & \textbf{75.00} / \textbf{65.00} & 20.00 / 40.00 \\
\hline
Natural Language Understanding & \textbf{62.5 / 100.00} & 25.00 / 0.00 & 62.5 / 50.00 \\
\hline
Use of World Knowledge & 16.67 / 66.67 & 33.33 / 0.00 & \textbf{100.00} / \textbf{83.33} \\
\hline
Multilingual Knowledge and Reasoning  & \textbf{100.00} / \textbf{100.00} & 0.00 / 0.00 & 50.00 / 50.00 \\
\hline
\rowcolor{gray!15}
Overall & 50.00 / \textcolor{blue}{\textbf{67.65}} & \textcolor{blue}{\textbf{55.88}} / 38.24 & 44.12 / 50.00 \\
\midrule
\rowcolor{orange!20}
\textbf{Claude 3 Sonnet $\rightarrow$ Claude 3 Haiku (Table ~\ref{table:cross_version_perf_haiku})} & & &\\
Algorithmic and Multi-Step Arithmetic Reasoning &  45.00 / 50.00 & \textbf{70.00} / 45.00 & 40.00 / \textbf{65.00}\\
\hline
Natural Language Understanding & 50.00 / 75.00 & 25.00 / 12.5 & \textbf{75.00} / \textbf{75.00} \\
\hline
Use of World Knowledge &  50.00 / 33.33 & \textbf{50.00} / \textbf{83.33} & 50.00 / 50.00 \\
\hline
Multilingual Knowledge and Reasoning  &  50.00 / 50.00 & 0.00 / 0.00 & \textbf{100.00} / \textbf{100.00}\\
\hline
\rowcolor{gray!15}
Overall & 44.12 / 52.94 & \textcolor{blue}{\textbf{55.88}} / 44.12 & 50.00 / \textcolor{blue}{\textbf{64.71}} \\
\midrule
\rowcolor{orange!20}
\textbf{Claude 3 Sonnet $\rightarrow$ LLAMA 3 (Table ~\ref{table:llama-adapatation}) } & & &\\
Algorithmic and Multi-Step Arithmetic Reasoning &  40.00 / \textbf{55.00} & \textbf{60.00} / 40.00 & 50.00 / 55.00 \\
\hline
Natural Language Understanding & 37.50 / 25.00 & 37.50 / \textbf{100.0}  & \textbf{75.00} / 25.00 \\
\hline
Use of World Knowledge & 50.00 / \textbf{83.33} & 16.67 / 50.00 & \textbf{83.33} / 16.67  \\
\hline
Multilingual Knowledge and Reasoning  &  50.00 / 50.00 & 0.00 / 100.00 & 100.00/ 0.00\\
\hline
\rowcolor{gray!15}
Overall & 41.18 / 52.94 & 47.06 / \textcolor{blue}{\textbf{55.88}} & \textcolor{blue}{\textbf{61.76}} / 41.18 \\
\hline
\end{tabular}
}
\end{center}
\end{table}

\subsubsection{Prompt Adaptation}

Next, we present the results of our experiments on adapting prompts across different model versions, model families, and languages from a source model/language to target model/language.

\textbf{Cross-model version and family.} Table~\ref{table:adaptation_summary} presents a summary of adapting prompts generated from Claude-3-haiku to the more advanced Claude-3-sonnet, and vice-versa. We also present cross-model family results with Claude-3-sonnet to LLAMA prompt adaptation. Detailed results can be found in Table~\ref{table:cross_version_perf_sonnet}, Table~\ref{table:cross_version_perf_haiku}, and Table~\ref{table:llama-adapatation} in appendix.
We compare LCP adaptation with directly performing the prompt optimization process on the target model from scratch (\textbf{LCP Optimization on Target}) and by using an optimized prompt (last iteration or best prompt on training set) from source model directly without any change. 

The results show that our adaptation framework effectively leverages feedback from the prior model version (even it is less effective) to enhance performance on the new model version --- it is typically better (best training prompt) or at par (last prompt) with prompt optimization from scratch on the target model. From the results in task types, one clear observation is how adaptation is a fine balance between the target model generated prompts (LCP Optimization on Target) and source model generated prompts (Source Optimized). For example, in Haiku $\rightarrow$ Sonnet setting, LCP Adaptation works better than source generated prompts but worse than target generated prompts for Algorithmic and multi-step Arithmetic reasoning tasks. Situation is completely reversed for Natural Language understanding tasks. This shows that prompt adaptation can slightly degrade performance compared to source on the tasks where the source model is relatively stronger while increasing the performance compared to source where the target model is relatively stronger. This observation is repeated even in the cross model setting. Hence, \emph{our LCP adaptation framework creates a balance between strengths of source and target models}. 

This observation can be attributed to our framework's ability to effectively leverage the strengths of the source model and transfer this knowledge to the prompts for target model via feedback. Our approach refines and tailors prompts to align with the nuances of the target model, complementing the target model. 
This is especially beneficial for scenarios where the tasks need target and source model's complementary capabilities making our approach a valuable tool that enables them to improve response quality even with weaker but specialized models, thereby expanding its applicability to a wider range of scenarios.

\textbf{Cross-lingual.} We report the results of our cross-lingual experiments in Table~\ref{table:cross-lingual}. We categorized the methods into two groups: \emph{prompt refinement} and \emph{query translation}. While our approach focuses on prompt refinement, we also present the results of query translation to provide additional insights. We compare our method with directly inputting the test query (Blank Prompt), adding an optimized English prompt generated by our prompt optimization method using the COPA dataset (Optimized Prompt), and translating the optimized prompt to the target language. The results indicate that our prompt adaptation approach outperforms the prompt baselines for 7 out of 11 languages. 

For query translation, we translate the input non-English language test query into English and either use a blank prompt or use the English prompt optimized by our method on the translated training data. Our results show that query translation works better than prompt refinement methods on 7 out of 11 tasks. This is in line with work from \cite{lin2022few}, where query translated worked better than human expert prompts in the query language. This is a function of English heavy training of current LLMs. It is important to note that query translation methods come with an additional computational cost, as each query must be translated into English before processing. However, as LLMs continue to improve their performance on non-English languages, we anticipate a narrowing of the gap between prompt refinement and query translation methods. Important to note that on two low resource languages: Swahili (\textbf{sw}) and Southern Quechua (\textbf{qu}), LCP even beats query translation methods. Our study not only presents a comprehensive analysis of cross-lingual performance but also introduces a novel prompt adaptation technique that bridges the gap between the prompt refinement and query translation methods.




\begin{table}[t]
\small
\caption{Cross-lingual accuracy on 11 languages in XCOPA dataset. We present the prompt with the best performance on training set. Performance numbers in \textcolor{blue}{\textbf{blue}} shows the best results for a language while in \textbf{bold} show best numbers among Prompt Refinement methods.}
\label{table:cross-lingual}
\begin{center}
\begin{tabular}{lccccccccccc}
\hline
& \multicolumn{11}{c}{\textbf{Language}} \\\cline{2-12}
\multicolumn{1}{c}{\bf Method} & \multicolumn{1}{c}{\bf ta}  &\multicolumn{1}{c}{\bf sw} &\multicolumn{1}{c}{\bf it} &\multicolumn{1}{c}{\bf ht} &\multicolumn{1}{c}{\bf et} &\multicolumn{1}{c}{\bf tr} &\multicolumn{1}{c}{\bf zh} &\multicolumn{1}{c}{\bf id} &\multicolumn{1}{c}{\bf qu} &\multicolumn{1}{c}{\bf th} &\multicolumn{1}{c}{\bf vi}\\
\hline
\rowcolor{orange!20}
\multicolumn{12}{c}{\textbf{Prompt Refinement}} \\
\hline
Blank Prompt & 79.6 & 79.0 & 94.8 & 79.0 & 89.4 & 90.4 & 90.8 & 94.0 & 59.8 & \textcolor{blue}{\textbf{88.2}} & \textbf{91.8}\\
Optimized Prompt & \textbf{82.4} & 79.4 & 91.4 & 80.0 & 91.6 & 90.6 & 88.8 & \textbf{94.2} & 54.0  & 86.0 & 91.6\\
Translated Prompt & 75.0 & 72.2 & 85.8 & 70.2 & 78.8 & 89.2 & 87.0 & 92.0 & 57.4 & 83.0 & 89.6\\
LCP &  80.8 &  \textbf{80.8} &  \textcolor{blue}{\textbf{96.8}} &  \textcolor{blue}{\textbf{83.2}} &  \textbf{92.2} & \textbf{92.6} &  \textbf{93.4} & 93.4 & \textcolor{blue}{\textbf{62.0}} & 85.8 & 91.2\\
\hline
\rowcolor{orange!20}
\multicolumn{12}{c}{\textbf{Query Translation}} \\
\hline
Blank Prompt & 89.6 & \textcolor{blue}{\textbf{83.6}} & 96.6 & 80.8 & \textcolor{blue}{\textbf{92.4}} & \textcolor{blue}{\textbf{94.2}} & 94.2 & \textcolor{blue}{\textbf{94.4}} & 61.4 & 86.6 & \textcolor{blue}{\textbf{92.2}}\\
Optimized Prompt & \textcolor{blue}{\textbf{91.4}} & 82.8 & 95.6 & 79.8 & 92.2 & 94.0 & \textcolor{blue}{\textbf{95.2}} & 91.0 & 58.2 & 86.2 & 90.8\\
\hline
\vspace{-1.5em}
\end{tabular}
\end{center}
\end{table}

\subsection{Ablation Study}
\label{sec:ablations}


\begin{figure}[h]
  \centering
    \includegraphics[width=\textwidth]{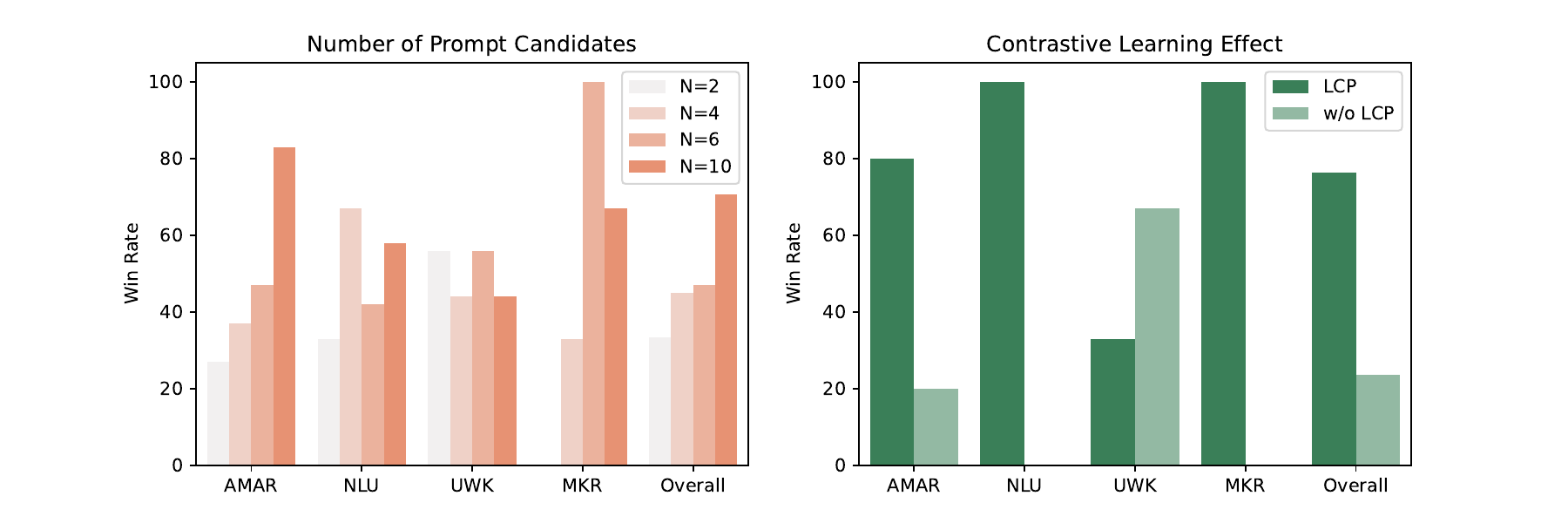}
  \caption{Ablation study with number of prompt candidates ($N$) on the left and Effect of contrastive learning (w/ and w/o constrastive learning) on the right. Reported are win rates on the prompt selected with best training set performance (best). AMAR refers to Algorithmic and Arithmetic, NLU to Natural Language Understanding, UWK to Understanding of World Knowledge, and MKR to Multilingual Knowledge and Reasoning categories.}
  \label{figure:lcp_ablations}
  \vspace{-1em}
\end{figure}


\paragraph{Diversity Injection with multiple prompt candidates}
We generate multiple prompt candidates ($N=10$) to explore the prompt space which is used for the our contrastive learning framework to identify the patterns between good and bad prompts from these prompt candidates evaluated on the training set. Multiple prompt candidates help us explore the diversity of the accuracy-prompt space, unlike previous methods dependent on single prompts. To explore the effectiveness of this mechanism, we use $N=2, 4$, and $6$, with top-$\floor{\frac{N}{2}}$ and bottom-$\floor{\frac{N}{2}}$ prompts used for contrastive learning. Win rates are shown in Figure~\ref{figure:lcp_ablations} and detailed results in Table~\ref{table:num_prompts_ablation}. We clearly see that the win rates increase as we increase $N$. This clearly demonstrates the effectiveness of injecting diversity while generating the prompt using contrastive learning with multiple prompt candidates. Increasing $N$ further, we saw limited benefit and a much higher computation cost, so we use $N=10$.

\paragraph{Contrastive Learning} One of the key contribution of our work is using contrastive learning to learn from both good and bad prompts, instead of just focusing on top prompts (OPRO) or wrong samples (AutoHint). We show the effectiveness of contrastive learning in our framework by comparing it with a setting providing the LLM with just the $N$ ranked prompt candidates similar to OPRO. Win rates are shown in Figure~\ref{figure:lcp_ablations} and detailed results in Table~\ref{table:contrastive_ablation}. Contrastive learning has a win rate of 76\%, especially benefiting the algorithmic and arithmetic tasks, which need more involved instruction writing. These results combined with diversity injection ablation clearly show the \emph{benefit of exploring and analyzing the prompt manifold to incorporate it in the prompt optimization}.

\begin{table}[h]
\small
\caption{Cross-optimizer performance comparison on various tasks for Claude-3-Haiku using Claude-3-Haiku (weaker version) versus Claude-3-Sonnet (stronger model) as prompt optimizer.}
\label{table:cross-optimizer}
\begin{center}
\begin{tabular}{lcc}
\hline
\multicolumn{1}{c}{\bf Task} & \multicolumn{2}{c}{\bf Optimizer}\\
& \multicolumn{1}{c}{\bf Haiku}  &\multicolumn{1}{c}{\bf Sonnet } \\
& Last/Best & Last/Best\\
\hline
date\_understanding & 24.0 / 69.0 & \bf 69.0 / 77.0 \\
reasoning\_about\_colored\_objects & 67.5 / 69.0 & \bf 70.0 / 71.0 \\
disambiguation\_qa & 61.0 / 63.5 & \bf 62.5 / 64.5 \\
logical\_deduction\_three\_objects & {\bf 70.0} / 68.0 & 69.0 / {\bf 71.5}\\
\hline
\end{tabular}
\end{center}
\end{table}

\textbf{Cross Optimizers.} Based on the results from prompt adaptation, a natural question arises: \emph{can we use stronger models to optimize prompts for a weaker model?} We aimed to investigate whether employing a more advanced model as the optimizer could further enhance performance. During the prompt optimization of Claude-3-Haiku, we utilize Claude-3-Sonnet to generate new candidate prompts, while still using Claude-3-Haiku for evaluation. We run this on selected four tasks due to cost constraints as shown in  Table~\ref{table:cross-optimizer}. We observe this approach significantly improves performance due to the capabilities of Claude-3-Sonnet. Claude-3-Sonnet as optimizer more effectively improved best/last prompts by 3.5\%/7\% on the four selected tasks. These results demonstrate the promising direction of leveraging more advanced models to optimize prompts for weaker models.

\begin{figure}[t!]
\centering
\begin{subfigure}{.3\textwidth}
  \centering
  \includegraphics[width=\linewidth]{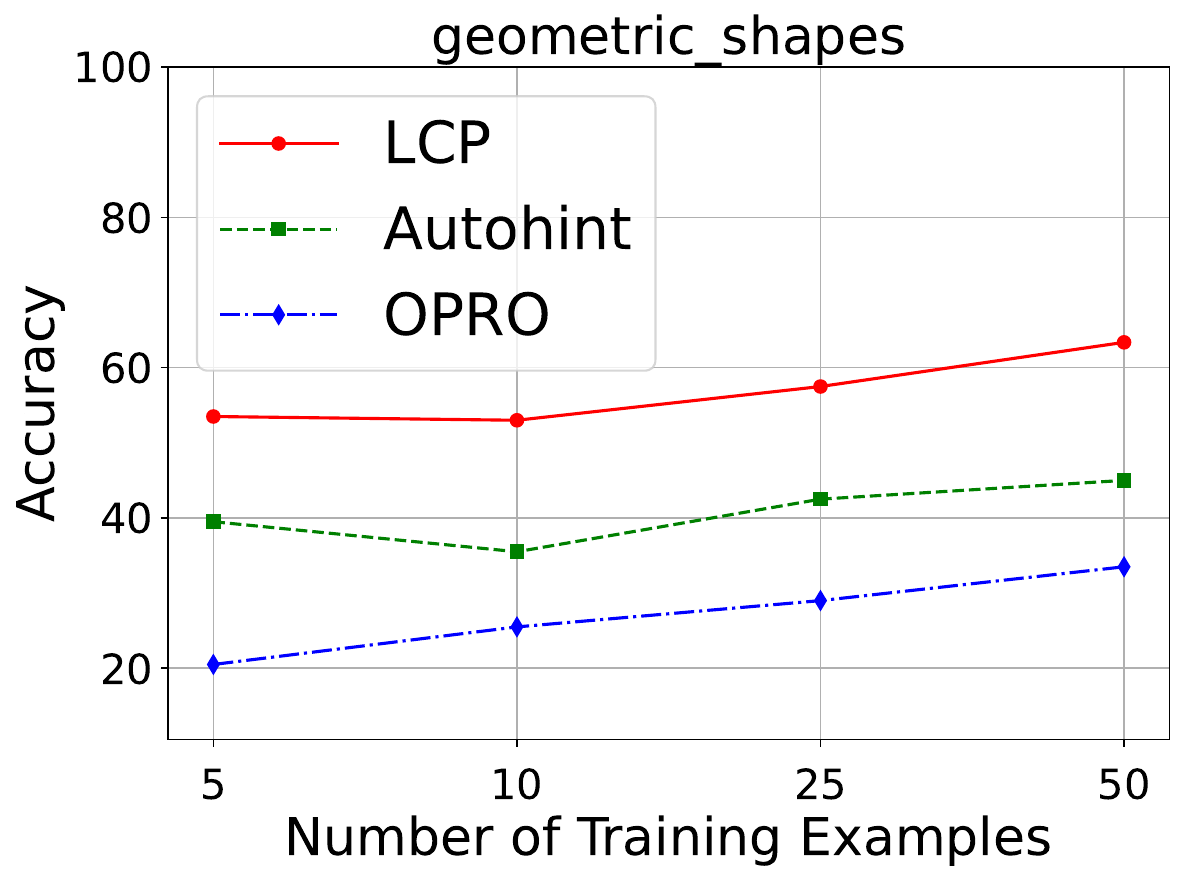}
  \caption{geometric\_shapes task}
  \label{figure:geometric_shapes}
\end{subfigure}%
\hfill
\begin{subfigure}{.3\textwidth}
  \centering
  \includegraphics[width=\linewidth]{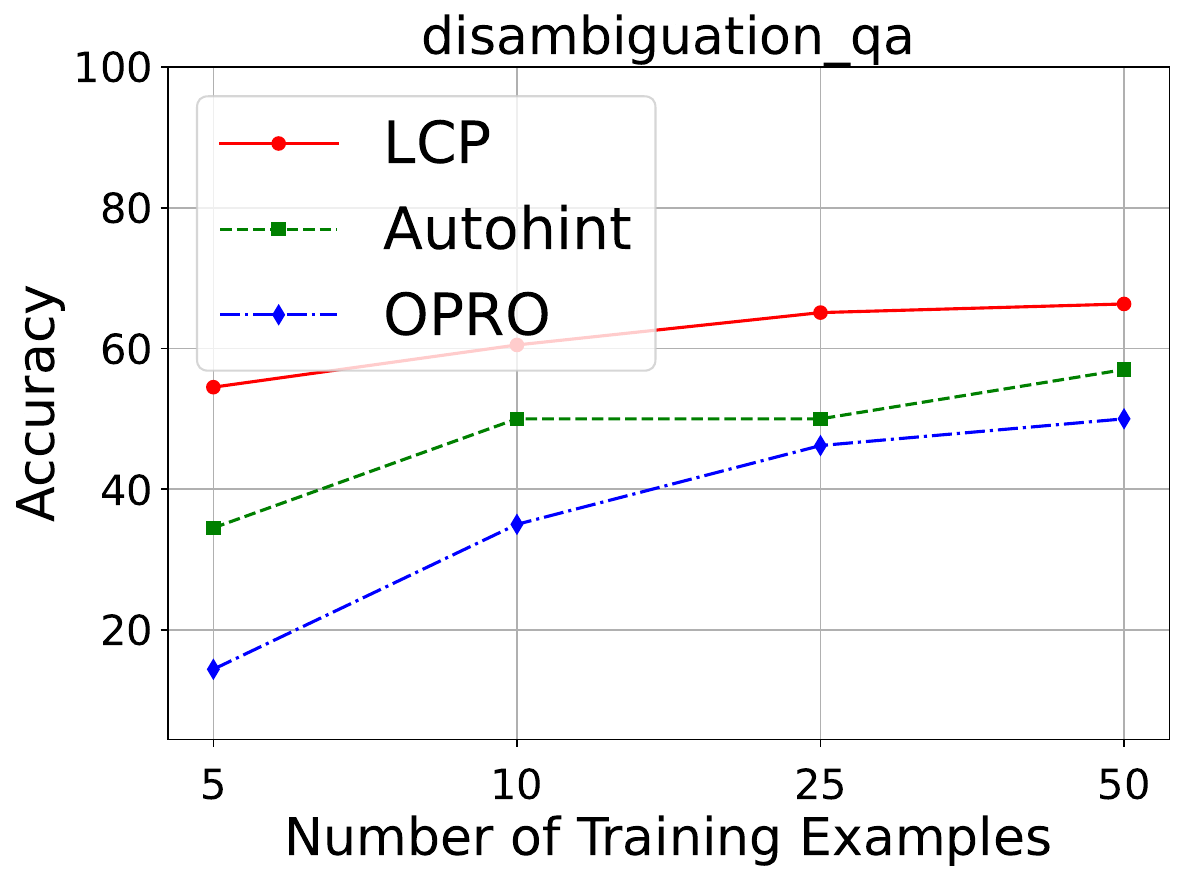}
  \caption{disambiguation\_qa task}
  \label{figure:disambiguation}
\end{subfigure}%
\hfill
\begin{subfigure}{.3\textwidth}
  \centering
  \includegraphics[width=\linewidth]{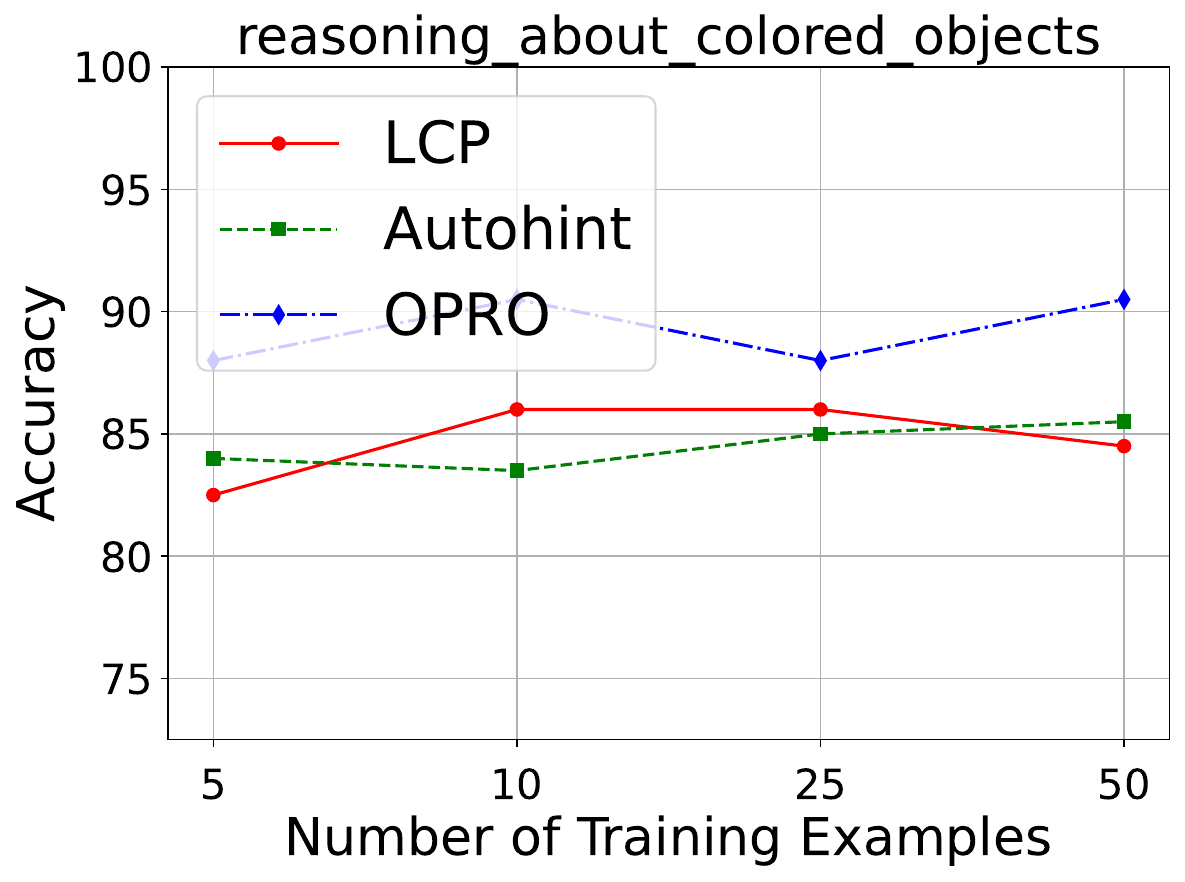}
  \caption{reasoning colored objects}
  \label{figure:date_understanding}
\end{subfigure}
\caption{Accuracy as a function of number of training examples for LCP, AutoHint, and OPRO.}
\label{figure:ablation_num_examples}
\end{figure}

\textbf{Number of Training Examples.} To provide insights into the number of examples required for our method to maintain effectiveness, we report the performance when using 5, 10, 25, and 50 examples for training, in Figure~\ref{figure:ablation_num_examples} for three tasks. We notice a trend when we analyzed the training plots. For tasks like \texttt{reasoning about colored objects} whose training accuracy was relatively flat during the iterations, number of examples had little effect, while for tasks like \texttt{geometric shapes} with training curves showing considerable improvement across training iterations, we see a consistent improvement in the performance as number of examples increased.  Further, we observe that LCP is relatively more sample efficient, giving a relatively higher performance at lower number of samples versus AutoHint or OPRO. This can be attributed to our multiple candidate generation for contrastive learning that helps model explore diverse prompts to derive insights.










\section{Related Work}

\textbf{Soft prompt optimization.} Recent studies have explored soft prompt-tuning, which involves prepending continuous vectors that are out of vocabulary and serve as prompts for specific tasks (\citealp{li-liang-2021-prefix}; \citealp{lester-etal-2021-power}; \citealp{liu-etal-2022-p}; \citealp{qin-eisner-2021-learning}; \citealp{ben2022pada}). These approaches are model-agnostic, but they do not focus on adapting prompts to other models. Some work does leverage soft prompt-tuning for cross-lingual adaptation (\citealp{li-etal-2023-enhancing-cross}; \citealp{huang-etal-2022-zero}; \citealp{zhao2023breaking}; \citealp{park2023analysis}). However, these methods often require access to model logits, which can be a constraint in practical applications with the recent proprietary models.

\textbf{Hard prompt optimization.} Hard prompt optimization involves crafting discrete, human-interpretable prompts. Prior works have focused on prompt engineering, iterative refinement, and search-based techniques to improve performance (\citealp{guo2024connecting}; \citealp{zhou2023large}; \citealp{pryzant-etal-2023-automatic}; \citealp{wang2024promptagent}; \citealp{sun2023autohint}; \citealp{yang2024large}; \citealp{wan-etal-2023-better}; \citealp{li-etal-2023-robust}; \citealp{yuksekgonul2024textgrad}). Specifically, APE~\citep{zhou2023large} selects the top instructions with the highest accuracies and then prompts the LLM to generate semantically similar variants for each selected instruction. AutoHint~\citep{sun2023autohint} generates hints from incorrect samples, summarizes these hints, and uses the summary as the new prompt. OPRO~\citep{yang2024large} generates new prompts by leveraging historical prompts and their corresponding scores, instructing the LLM to create improved versions. 

While these works focused on prompt optimization they did not explore prompt adaptability to various model versions, model families, and languages. Our proposed approach bridges this gap by providing a novel comprehensive framework for prompt optimization and adaptation, ensuring effectiveness across different models and linguistic contexts using contrastive learning.

\section{Conclusion}

In this paper, we proposed Learning from Contrastive Prompts (LCP), a comprehensive framework for prompt optimization and adaptation. Our approach addresses the limitations of existing optimization methods and addresses an unexplored but common problem of adapting prompts across different model versions, model families, and languages. It involves a systematic process of prompt candidate generation and new prompt generation through contrastive learning, and feedback for prompt adaptation setting to ensure that the prompts remain effective and relevant in diverse scenarios. We conducted extensive experiments on the Big-Bench Hard dataset, demonstrating that our framework significantly outperforms existing methods. 

Our results showed that our approach maintains high performance when adapting prompts across different model versions complementing the strengths of the source and target models. Additionally, our framework proved robust in cross-lingual scenarios, effectively handling the challenges posed by different linguistic contexts. Our results also show that using a stronger model for prompt optimization and adaptation could significantly boost performance on weaker LLMs instead of prompt adaptation from scratch using our framework.




One of the key areas of investigation from our work is exploration and exploiting prompt manifold in a more systematic way. The current prompt optimization methods including ours are unstable over iterations, and its not clear how to navigate the prompt manifold (see Appendix~\ref{sec:prompt_similarity} for more discussion). Some avenues could include a richer feedback mechanism across iterations, as we only rely on feedback from the prompt generated in the preceding iteration or giving a higher weightage to better hints. Further, letting the LLMs explain the feedback and incorporating that reasoning could also be potentially helpful. Prompt adaptation, which can be thought of through the lens of classical domain adaptation can be helped by a more sophisticated feedback design to get best of both the target model and source model, as we see it currently tries to strike the balance between the two. Our cross-optimizer results also show a promising direction and needs further exploration, especially in domains like law or medical where weaker but domain specialized model could guide or be guided in a collaborative fashion by more powerful general LLMs to generate powerful prompts. 

\newpage

\bibliography{paper}

\begin{thebibliography}{29}
\providecommand{\natexlab}[1]{#1}
\providecommand{\url}[1]{\texttt{#1}}
\expandafter\ifx\csname urlstyle\endcsname\relax
  \providecommand{\doi}[1]{doi: #1}\else
  \providecommand{\doi}{doi: \begingroup \urlstyle{rm}\Url}\fi

\bibitem[Anthropic(2024)]{anthropic2024claude}
AI~Anthropic.
\newblock The claude 3 model family: Opus, sonnet, haiku.
\newblock \emph{Claude-3 Model Card}, 1, 2024.

\bibitem[Ben-David et~al.(2022)Ben-David, Oved, and Reichart]{ben2022pada}
Eyal Ben-David, Nadav Oved, and Roi Reichart.
\newblock Pada: Example-based prompt learning for on-the-fly adaptation to
  unseen domains.
\newblock \emph{Transactions of the Association for Computational Linguistics},
  10:\penalty0 414--433, 2022.

\bibitem[Chen et~al.(2020)Chen, Kornblith, Norouzi, and Hinton]{chen2020simple}
Ting Chen, Simon Kornblith, Mohammad Norouzi, and Geoffrey Hinton.
\newblock A simple framework for contrastive learning of visual
  representations.
\newblock In \emph{International conference on machine learning}, pp.\
  1597--1607. PMLR, 2020.

\bibitem[Dubey et~al.(2024)Dubey, Jauhri, Pandey, Kadian, Al-Dahle, Letman,
  Mathur, Schelten, Yang, Fan, et~al.]{dubey2024llama}
Abhimanyu Dubey, Abhinav Jauhri, Abhinav Pandey, Abhishek Kadian, Ahmad
  Al-Dahle, Aiesha Letman, Akhil Mathur, Alan Schelten, Amy Yang, Angela Fan,
  et~al.
\newblock The llama 3 herd of models.
\newblock \emph{arXiv preprint arXiv:2407.21783}, 2024.

\bibitem[Guo et~al.(2024)Guo, Wang, Guo, Li, Song, Tan, Liu, Bian, and
  Yang]{guo2024connecting}
Qingyan Guo, Rui Wang, Junliang Guo, Bei Li, Kaitao Song, Xu~Tan, Guoqing Liu,
  Jiang Bian, and Yujiu Yang.
\newblock Connecting large language models with evolutionary algorithms yields
  powerful prompt optimizers.
\newblock In \emph{The Twelfth International Conference on Learning
  Representations}, 2024.
\newblock URL \url{https://openreview.net/forum?id=ZG3RaNIsO8}.

\bibitem[Huang et~al.(2022)Huang, Ma, Zhang, Wei, and
  Wang]{huang-etal-2022-zero}
Lianzhe Huang, Shuming Ma, Dongdong Zhang, Furu Wei, and Houfeng Wang.
\newblock Zero-shot cross-lingual transfer of prompt-based tuning with a
  unified multilingual prompt.
\newblock In Yoav Goldberg, Zornitsa Kozareva, and Yue Zhang (eds.),
  \emph{Proceedings of the 2022 Conference on Empirical Methods in Natural
  Language Processing}, pp.\  11488--11497, Abu Dhabi, United Arab Emirates,
  December 2022. Association for Computational Linguistics.
\newblock \doi{10.18653/v1/2022.emnlp-main.790}.
\newblock URL \url{https://aclanthology.org/2022.emnlp-main.790}.

\bibitem[Kojima et~al.(2022)Kojima, Gu, Reid, Matsuo, and
  Iwasawa]{kojima2022large}
Takeshi Kojima, Shixiang~Shane Gu, Machel Reid, Yutaka Matsuo, and Yusuke
  Iwasawa.
\newblock Large language models are zero-shot reasoners.
\newblock \emph{Advances in neural information processing systems},
  35:\penalty0 22199--22213, 2022.

\bibitem[Lester et~al.(2021)Lester, Al-Rfou, and
  Constant]{lester-etal-2021-power}
Brian Lester, Rami Al-Rfou, and Noah Constant.
\newblock The power of scale for parameter-efficient prompt tuning.
\newblock In Marie-Francine Moens, Xuanjing Huang, Lucia Specia, and Scott
  Wen-tau Yih (eds.), \emph{Proceedings of the 2021 Conference on Empirical
  Methods in Natural Language Processing}, pp.\  3045--3059, Online and Punta
  Cana, Dominican Republic, November 2021. Association for Computational
  Linguistics.
\newblock \doi{10.18653/v1/2021.emnlp-main.243}.
\newblock URL \url{https://aclanthology.org/2021.emnlp-main.243}.

\bibitem[Li et~al.(2023{\natexlab{a}})Li, Wang, Feng, Cao, Zhang, and
  Chua]{li-etal-2023-robust}
Moxin Li, Wenjie Wang, Fuli Feng, Yixin Cao, Jizhi Zhang, and Tat-Seng Chua.
\newblock Robust prompt optimization for large language models against
  distribution shifts.
\newblock In Houda Bouamor, Juan Pino, and Kalika Bali (eds.),
  \emph{Proceedings of the 2023 Conference on Empirical Methods in Natural
  Language Processing}, pp.\  1539--1554, Singapore, December
  2023{\natexlab{a}}. Association for Computational Linguistics.
\newblock \doi{10.18653/v1/2023.emnlp-main.95}.
\newblock URL \url{https://aclanthology.org/2023.emnlp-main.95}.

\bibitem[Li et~al.(2023{\natexlab{b}})Li, Hu, Liu, Yang, Ma, Yu, and
  Wen]{li-etal-2023-enhancing-cross}
Shuang Li, Xuming Hu, Aiwei Liu, Yawen Yang, Fukun Ma, Philip~S. Yu, and Lijie
  Wen.
\newblock Enhancing cross-lingual natural language inference by soft prompting
  with multilingual verbalizer.
\newblock In Anna Rogers, Jordan Boyd-Graber, and Naoaki Okazaki (eds.),
  \emph{Findings of the Association for Computational Linguistics: ACL 2023},
  pp.\  1361--1374, Toronto, Canada, July 2023{\natexlab{b}}. Association for
  Computational Linguistics.
\newblock \doi{10.18653/v1/2023.findings-acl.88}.
\newblock URL \url{https://aclanthology.org/2023.findings-acl.88}.

\bibitem[Li \& Liang(2021)Li and Liang]{li-liang-2021-prefix}
Xiang~Lisa Li and Percy Liang.
\newblock Prefix-tuning: Optimizing continuous prompts for generation.
\newblock In Chengqing Zong, Fei Xia, Wenjie Li, and Roberto Navigli (eds.),
  \emph{Proceedings of the 59th Annual Meeting of the Association for
  Computational Linguistics and the 11th International Joint Conference on
  Natural Language Processing (Volume 1: Long Papers)}, pp.\  4582--4597,
  Online, August 2021. Association for Computational Linguistics.
\newblock \doi{10.18653/v1/2021.acl-long.353}.
\newblock URL \url{https://aclanthology.org/2021.acl-long.353}.

\bibitem[Liang et~al.(2022)Liang, Bommasani, Lee, Tsipras, Soylu, Yasunaga,
  Zhang, Narayanan, Wu, Kumar, et~al.]{liang2022holistic}
Percy Liang, Rishi Bommasani, Tony Lee, Dimitris Tsipras, Dilara Soylu,
  Michihiro Yasunaga, Yian Zhang, Deepak Narayanan, Yuhuai Wu, Ananya Kumar,
  et~al.
\newblock Holistic evaluation of language models.
\newblock \emph{arXiv preprint arXiv:2211.09110}, 2022.

\bibitem[Lin et~al.(2022)Lin, Mihaylov, Artetxe, Wang, Chen, Simig, Ott, Goyal,
  Bhosale, Du, et~al.]{lin2022few}
Xi~Victoria Lin, Todor Mihaylov, Mikel Artetxe, Tianlu Wang, Shuohui Chen,
  Daniel Simig, Myle Ott, Naman Goyal, Shruti Bhosale, Jingfei Du, et~al.
\newblock Few-shot learning with multilingual generative language models.
\newblock In \emph{Proceedings of the 2022 Conference on Empirical Methods in
  Natural Language Processing}, pp.\  9019--9052, 2022.

\bibitem[Liu et~al.(2022)Liu, Ji, Fu, Tam, Du, Yang, and Tang]{liu-etal-2022-p}
Xiao Liu, Kaixuan Ji, Yicheng Fu, Weng Tam, Zhengxiao Du, Zhilin Yang, and Jie
  Tang.
\newblock {P}-tuning: Prompt tuning can be comparable to fine-tuning across
  scales and tasks.
\newblock In Smaranda Muresan, Preslav Nakov, and Aline Villavicencio (eds.),
  \emph{Proceedings of the 60th Annual Meeting of the Association for
  Computational Linguistics (Volume 2: Short Papers)}, pp.\  61--68, Dublin,
  Ireland, May 2022. Association for Computational Linguistics.
\newblock \doi{10.18653/v1/2022.acl-short.8}.
\newblock URL \url{https://aclanthology.org/2022.acl-short.8}.

\bibitem[Park et~al.(2023)Park, Park, Yoo, and Yoon]{park2023analysis}
Nohil Park, Joonsuk Park, Kang~Min Yoo, and Sungroh Yoon.
\newblock On the analysis of cross-lingual prompt tuning for decoder-based
  multilingual model.
\newblock \emph{arXiv preprint arXiv:2311.07820}, 2023.

\bibitem[Ponti et~al.(2020)Ponti, Glava{\v{s}}, Majewska, Liu, Vuli{\'c}, and
  Korhonen]{ponti-etal-2020-xcopa}
Edoardo~Maria Ponti, Goran Glava{\v{s}}, Olga Majewska, Qianchu Liu, Ivan
  Vuli{\'c}, and Anna Korhonen.
\newblock {XCOPA}: A multilingual dataset for causal commonsense reasoning.
\newblock In Bonnie Webber, Trevor Cohn, Yulan He, and Yang Liu (eds.),
  \emph{Proceedings of the 2020 Conference on Empirical Methods in Natural
  Language Processing (EMNLP)}, pp.\  2362--2376, Online, November 2020.
  Association for Computational Linguistics.
\newblock \doi{10.18653/v1/2020.emnlp-main.185}.
\newblock URL \url{https://aclanthology.org/2020.emnlp-main.185}.

\bibitem[Pryzant et~al.(2023)Pryzant, Iter, Li, Lee, Zhu, and
  Zeng]{pryzant-etal-2023-automatic}
Reid Pryzant, Dan Iter, Jerry Li, Yin Lee, Chenguang Zhu, and Michael Zeng.
\newblock Automatic prompt optimization with {``}gradient descent{''} and beam
  search.
\newblock In Houda Bouamor, Juan Pino, and Kalika Bali (eds.),
  \emph{Proceedings of the 2023 Conference on Empirical Methods in Natural
  Language Processing}, pp.\  7957--7968, Singapore, December 2023. Association
  for Computational Linguistics.
\newblock \doi{10.18653/v1/2023.emnlp-main.494}.
\newblock URL \url{https://aclanthology.org/2023.emnlp-main.494}.

\bibitem[Qin \& Eisner(2021)Qin and Eisner]{qin-eisner-2021-learning}
Guanghui Qin and Jason Eisner.
\newblock Learning how to ask: Querying {LM}s with mixtures of soft prompts.
\newblock In Kristina Toutanova, Anna Rumshisky, Luke Zettlemoyer, Dilek
  Hakkani-Tur, Iz~Beltagy, Steven Bethard, Ryan Cotterell, Tanmoy Chakraborty,
  and Yichao Zhou (eds.), \emph{Proceedings of the 2021 Conference of the North
  American Chapter of the Association for Computational Linguistics: Human
  Language Technologies}, pp.\  5203--5212, Online, June 2021. Association for
  Computational Linguistics.
\newblock \doi{10.18653/v1/2021.naacl-main.410}.
\newblock URL \url{https://aclanthology.org/2021.naacl-main.410}.

\bibitem[Rafailov et~al.(2024)Rafailov, Sharma, Mitchell, Manning, Ermon, and
  Finn]{rafailov2024direct}
Rafael Rafailov, Archit Sharma, Eric Mitchell, Christopher~D Manning, Stefano
  Ermon, and Chelsea Finn.
\newblock Direct preference optimization: Your language model is secretly a
  reward model.
\newblock \emph{Advances in Neural Information Processing Systems}, 36, 2024.

\bibitem[Reimers \& Gurevych(2019)Reimers and Gurevych]{reimers2019sentence}
Nils Reimers and Iryna Gurevych.
\newblock Sentence-bert: Sentence embeddings using siamese bert-networks.
\newblock In \emph{Proceedings of the 2019 Conference on Empirical Methods in
  Natural Language Processing and the 9th International Joint Conference on
  Natural Language Processing (EMNLP-IJCNLP)}, pp.\  3982--3992, 2019.

\bibitem[Salinas \& Morstatter(2024)Salinas and
  Morstatter]{salinas2024butterfly}
Abel Salinas and Fred Morstatter.
\newblock The butterfly effect of altering prompts: How small changes and
  jailbreaks affect large language model performance.
\newblock \emph{arXiv preprint arXiv:2401.03729}, 2024.

\bibitem[Sun et~al.(2023)Sun, Li, Xu, Homma, Cao, Wu, Jiao, and
  Charles]{sun2023autohint}
Hong Sun, Xue Li, Yinchuan Xu, Youkow Homma, Qi~Cao, Min Wu, Jian Jiao, and
  Denis Charles.
\newblock Autohint: Automatic prompt optimization with hint generation.
\newblock \emph{arXiv preprint arXiv:2307.07415}, 2023.

\bibitem[Suzgun et~al.(2022)Suzgun, Scales, Sch{\"a}rli, Gehrmann, Tay, Chung,
  Chowdhery, Le, Chi, Zhou, et~al.]{suzgun2022challenging}
Mirac Suzgun, Nathan Scales, Nathanael Sch{\"a}rli, Sebastian Gehrmann, Yi~Tay,
  Hyung~Won Chung, Aakanksha Chowdhery, Quoc~V Le, Ed~H Chi, Denny Zhou, et~al.
\newblock Challenging big-bench tasks and whether chain-of-thought can solve
  them.
\newblock \emph{arXiv preprint arXiv:2210.09261}, 2022.

\bibitem[Wan et~al.(2023)Wan, Sun, Dai, Arik, and
  Pfister]{wan-etal-2023-better}
Xingchen Wan, Ruoxi Sun, Hanjun Dai, Sercan Arik, and Tomas Pfister.
\newblock Better zero-shot reasoning with self-adaptive prompting.
\newblock In Anna Rogers, Jordan Boyd-Graber, and Naoaki Okazaki (eds.),
  \emph{Findings of the Association for Computational Linguistics: ACL 2023},
  pp.\  3493--3514, Toronto, Canada, July 2023. Association for Computational
  Linguistics.
\newblock \doi{10.18653/v1/2023.findings-acl.216}.
\newblock URL \url{https://aclanthology.org/2023.findings-acl.216}.

\bibitem[Wang et~al.(2024)Wang, Li, Wang, Bai, Luo, Zhang, Jojic, Xing, and
  Hu]{wang2024promptagent}
Xinyuan Wang, Chenxi Li, Zhen Wang, Fan Bai, Haotian Luo, Jiayou Zhang, Nebojsa
  Jojic, Eric Xing, and Zhiting Hu.
\newblock Promptagent: Strategic planning with language models enables
  expert-level prompt optimization.
\newblock In \emph{The Twelfth International Conference on Learning
  Representations}, 2024.
\newblock URL \url{https://openreview.net/forum?id=22pyNMuIoa}.

\bibitem[Yang et~al.(2024)Yang, Wang, Lu, Liu, Le, Zhou, and
  Chen]{yang2024large}
Chengrun Yang, Xuezhi Wang, Yifeng Lu, Hanxiao Liu, Quoc~V Le, Denny Zhou, and
  Xinyun Chen.
\newblock Large language models as optimizers.
\newblock In \emph{The Twelfth International Conference on Learning
  Representations}, 2024.
\newblock URL \url{https://openreview.net/forum?id=Bb4VGOWELI}.

\bibitem[Yuksekgonul et~al.(2024)Yuksekgonul, Bianchi, Boen, Liu, Huang,
  Guestrin, and Zou]{yuksekgonul2024textgrad}
Mert Yuksekgonul, Federico Bianchi, Joseph Boen, Sheng Liu, Zhi Huang, Carlos
  Guestrin, and James Zou.
\newblock Textgrad: Automatic" differentiation" via text.
\newblock \emph{arXiv preprint arXiv:2406.07496}, 2024.

\bibitem[Zhao et~al.(2023)Zhao, Chen, Lee, Qiu, Gao, Fan, and
  Lane]{zhao2023breaking}
Wanru Zhao, Yihong Chen, Royson Lee, Xinchi Qiu, Yan Gao, Hongxiang Fan, and
  Nicholas~Donald Lane.
\newblock Breaking physical and linguistic borders: Multilingual federated
  prompt tuning for low-resource languages.
\newblock In \emph{International Workshop on Federated Learning in the Age of
  Foundation Models in Conjunction with NeurIPS 2023}, 2023.
\newblock URL \url{https://openreview.net/forum?id=HyRwexERAo}.

\bibitem[Zhou et~al.(2023)Zhou, Muresanu, Han, Paster, Pitis, Chan, and
  Ba]{zhou2023large}
Yongchao Zhou, Andrei~Ioan Muresanu, Ziwen Han, Keiran Paster, Silviu Pitis,
  Harris Chan, and Jimmy Ba.
\newblock Large language models are human-level prompt engineers.
\newblock In \emph{The Eleventh International Conference on Learning
  Representations}, 2023.
\newblock URL \url{https://openreview.net/forum?id=92gvk82DE-}.

\end{thebibliography}
\bibliographystyle{iclr2025_conference}

\newpage
\appendix
\section{Appendix}

\subsection{Discussion about Validation Set}
\label{appendix:valid}

\begin{figure}[h]
\centering
\begin{subfigure}{.45\textwidth}
  \centering
  \includegraphics[width=\linewidth]{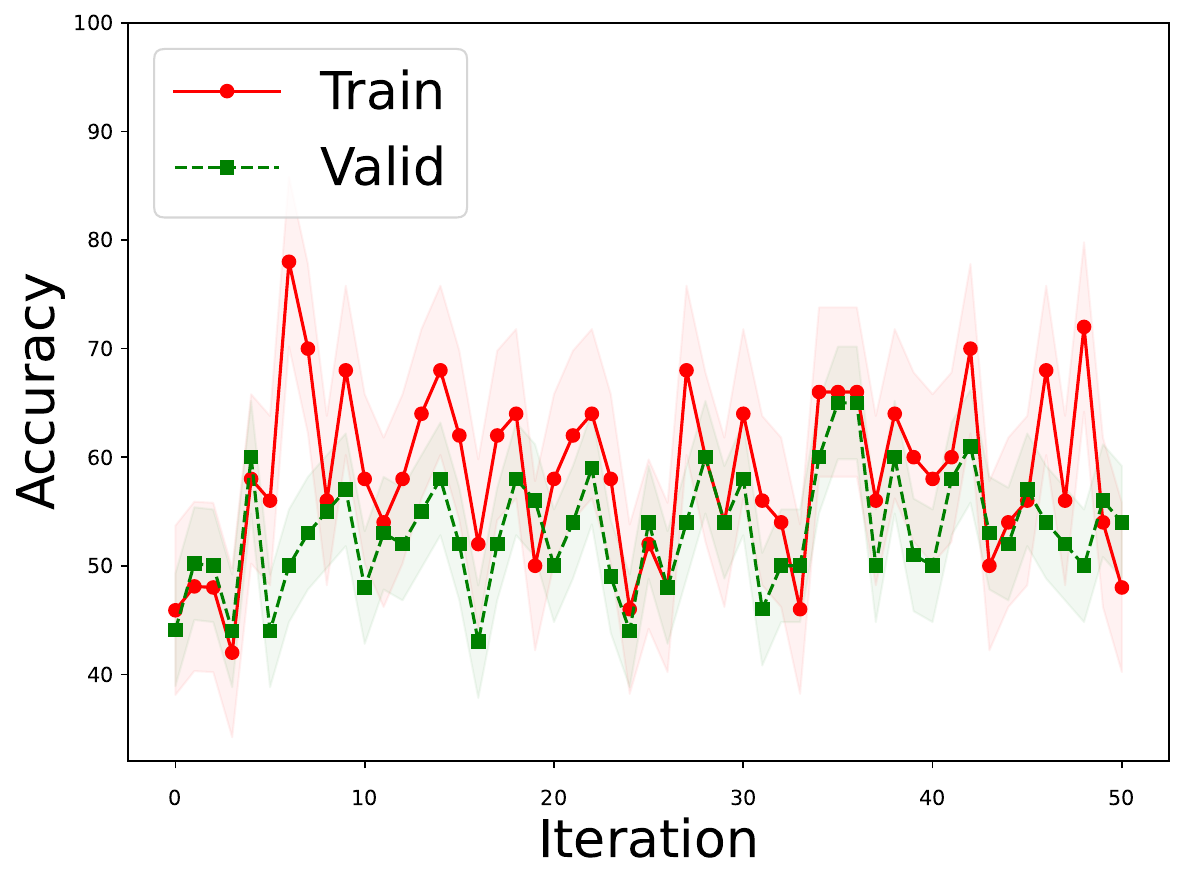}
  \caption{geometric\_shapes task}
  \label{figure:geometric_shapes_fit}
\end{subfigure}%
\hfill
\begin{subfigure}{.45\textwidth}
  \centering
  \includegraphics[width=\linewidth]{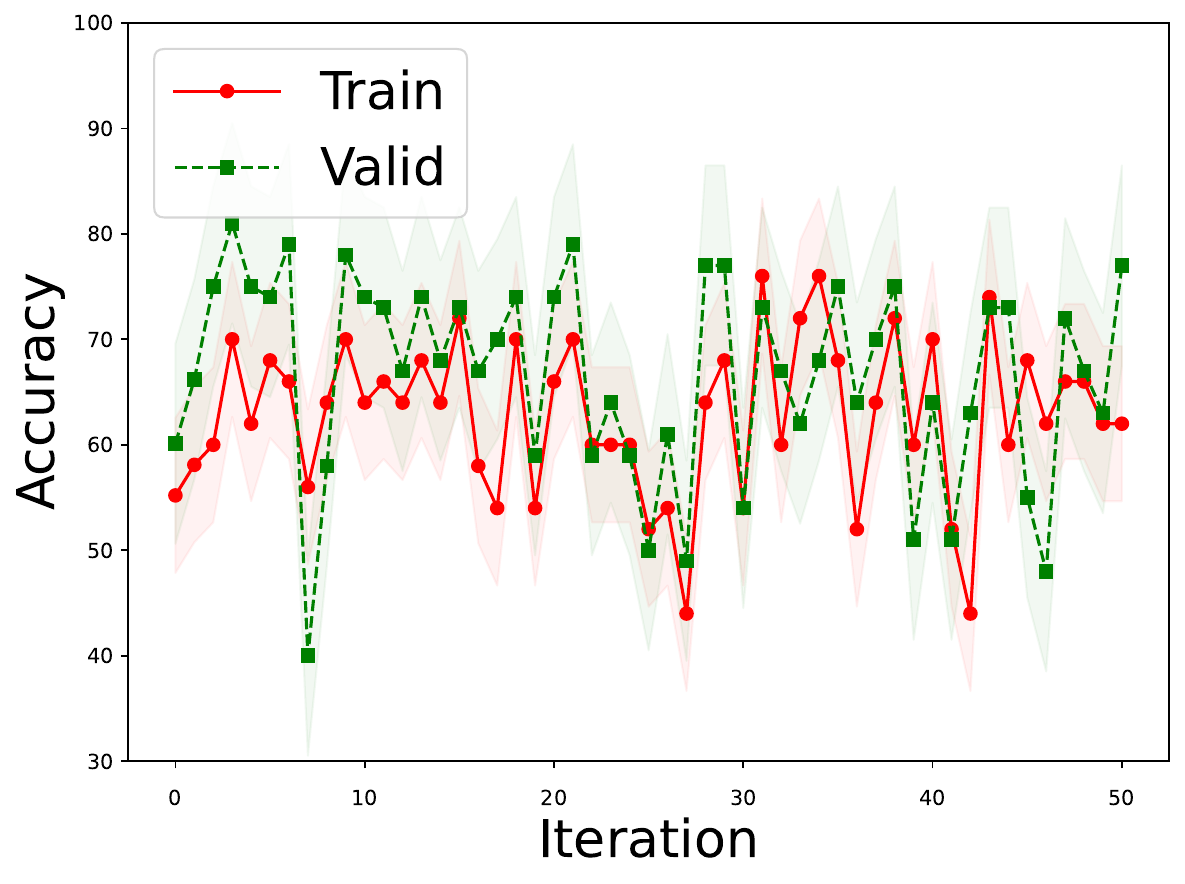}
  \caption{disambiguation\_qa task}
  \label{figure:disambiguation_fit}
\end{subfigure}%
\caption{Accuracy as a function of number of training examples for LCP, AutoHint, and OPRO.}
\label{figure:valid_set}
\end{figure}

We did not use a validation set for our experiments following OPRO~\citep{yang2024large} and AutoHint~\citep{sun2023autohint}, and based on our experiments. We show the training and validation accuracy curves when we do setup a validation set aside in Figure~\ref{figure:valid_set} on two tasks. We use a split of 33.33\% training, 33.33\% validation, and 33.33\% testing sets to show these results.

We observe that there is no inherent bias-variance trade-off between the training/validation accuracies; typically validation accuracy follows training accuracy. We observe a moderate Spearman's correlation between 0.45-0.55 (p-values$<$0.001), showing that they are quite correlated. Further, we see that our method does not really overfit; training accuracy is lower or similar to validation accuracy unlike overfitting exhibited by OPRO as noted by~\cite{yang2024large}.  Unlike traditional fine-tuning machine learning regime where the training data gets embedded into the model weights, it is quite clear on how to define overfitting in prompt optimization except prompts becoming too specific to the training samples. Since prompt accuracies change significantly iteration-over-iteration, further exploration is needed in this space to devise a way of final prompt selection. To keep it consistent with prior works~\citep{yang2024large, sun2023autohint} and to keep things simple in absence of an evidence of over-fitting, we chose to use last iteration prompt and prompt with best accuracy on training set.

\subsection{Prompt Similarity Visualization}
\label{sec:prompt_similarity}

\begin{figure}[h]
  \centering
   \includegraphics[width=7cm]{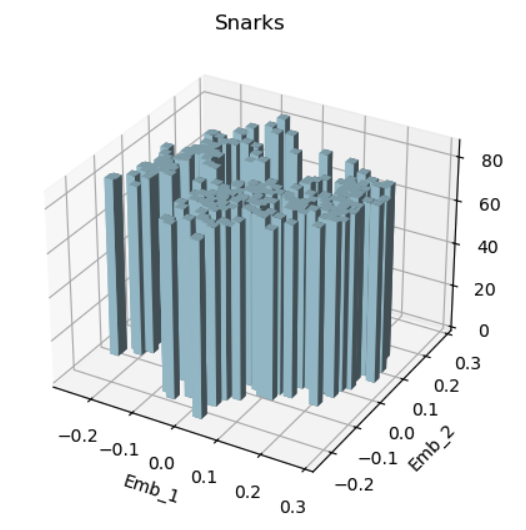}
 \caption{Visualization of prompt similarity: generated prompts in a 2D embedding space versus the performance on snarks task on the z-axis.}
 \label{figure:prompt_similarity}
\end{figure}

 We visualize the performance over prompt embeddings in Figure~\ref{figure:prompt_similarity}. Using sentence transformer~\citep{reimers2019sentence}, we embed the prompts generated over 50 iterations on the snarks task. This three-dimensional histogram plots the distribution of prompts in a two-dimensional embedding space selected using the first two principal components representing a reduced-dimensional representation of the prompt space.
The z-axis, represents the performance metric of each prompt. The varying heights of the uniform blue-gray bars illustrate the performance landscape across the embedding space.

Two regions appear prominently on this: low performing prompts facing us and high performing prompts facing away from us. There is a notion of a boundary line dividing the two regions. 
Our analysis of this visualization reveals that semantically similar prompts, represented by nearby points in the embedding space, tend to yield comparable performance results. This is evidenced by clusters of bars with similar heights. However, even slight changes in the prompt, especially for the prompts closer to the boundary, can lead to significant variations in performance, highlighting the sensitivity of optimization methods to prompt formulation.

It demonstrates that while semantic similarity often correlates with performance similarity, the relationship is not always straightforward. The complex landscape depicted here emphasizes the challenges and opportunities in prompt optimization with a hard to map prompt-accuracy manifold. Our diversity injection and contrastive learning framework helped explore and guide the prompt optimization through this space. More work needs to be done to understand how to create methods to navigate this manifold.

\subsection{Number of Contrastive Prompts} 
\begin{table}[h]
\small
\caption{Performance results with ablation on number of prompts used for Contrastive learning. We present the last / best performance for each task.}
\label{table:num_of_prompts}
\begin{center}
\begin{tabular}{lcccc}
\hline
\multicolumn{1}{c}{\bf Task} & \multicolumn{1}{c}{\bf 2}  &\multicolumn{1}{c}{\bf 3} &\multicolumn{1}{c}{\bf 4} &\multicolumn{1}{c}{\bf 5}\\
\hline
date\_understanding & \textbf{79.0} / 76.5 & 75.5 / 74.5 & 77.5 / 73.5 & 78.5 / \textbf{78.0}\\
reasoning\_about\_colored\_objects & \textbf{85.5} / 83.5 & 85.3 / \textbf{84.4} & 84.5 / 81.0 & 85.5 / 84.0 \\
\hline
\end{tabular}
\end{center}
\end{table}

Table~\ref{table:num_of_prompts} shows the ablation of number of selected prompts for contrastive learning's feedback. We observe there is some variation in the performance across the number of prompts but no clear trend. Hence, no clear choice of which number of prompts to select. We chose 3 as higher number of prompts incur much more computation costs.

\subsection{Number of selected wrong data samples} AutoHint~\citep{sun2023autohint} observed that using no more than 3 samples per iteration achieves the best performance, as more samples could confuse the LLM when generating the summary. We also investigate how the performance varies with different numbers of selected wrong samples in Table~\ref{table:num_of_wrong_samples}. We do not observe a clear benefit of increasing the number from three. Hence, we use three wrong samples during our experiments in accordance with AutoHint.

\begin{table}[h]
\small
\caption{Performance results with ablation on number of wrong samples performance. We present the last / best performance for each task.}
\label{table:num_of_wrong_samples}
\begin{center}
\begin{tabular}{lcccc}
\hline
\multicolumn{1}{c}{\bf Task} & \multicolumn{1}{c}{\bf 3}  &\multicolumn{1}{c}{\bf 4} &\multicolumn{1}{c}{\bf 5} &\multicolumn{1}{c}{\bf 6}\\
\hline
date\_understanding & 75.5 / 74.5  & 70.5 / \bf{75.0} & \textbf{77.0} / 72.5 & \textbf{77.0} / 74.0\\
reasoning\_about\_colored\_objects & \textbf{85.3} / 84.4 & 81.0 / \bf{85.5} & 79.5 / 82.5 & 82.5 / 83.0 \\
\hline
\end{tabular}
\end{center}
\end{table}

\subsection{Meta prompt}
\label{sec:meta_prompts}

\begin{tcolorbox}[colback=white, colframe=black, colbacktitle=white, coltitle=orange, boxrule=0.2mm]
\vspace{-8pt}
\tcbsubtitle{\textbf{Reason Generation Prompt}}
Given input: [INPUT]

And its expected output: [OUTPUT]

\vspace{5pt}

Explain the reason why the input corresponds to the given expected output. The reason should be placed within tag $<$reason$>$$<$/reason$>$. 

\tcbsubtitle{\textbf{Summarization Prompt}}

Given input and expected output pairs, along with the reason for generated outputs, provide a summarized common reason applicable to all cases within tags $<$summary$>$ and $<$/summary$>$.

The summary should explain the underlying principles, logic or methodology governing the relationship between the inputs and corresponding outputs. Avoid mentioning any specific details, numbers, or entities from the individual examples, and aim for a generalized explanation. 

\tcbsubtitle{\textbf{High-level Contrastive Prompt}}
Given $m$ examples of good prompts and their corresponding scores and $m$ examples of bad prompts and their corresponding scores, explore the unerlying pattern of good prompts, generate a new prompt based on this pattern. Put the new prompt within tag $<$prompt$>$ and $<$/prompt$>$. 

\vspace{5pt}

Good prompts and scores:

Prompt 1: [PROMPT 1]

Score: [SCORE 1]

...

Prompt $m$: [PROMPT $m$]

Score: [SCORE $m$]

\tcbsubtitle[center]{\textbf{Low-level Contrastive Prompts}}

Given $m$ prompt pairs and their corresponding scores, explain why one prompt is better than others. 

\vspace{5pt}
Prompt pairs and scores:

Prompt 1: [PROMPT 1]

Score: [SCORE 1]

...

Prompt $m$: [PROMPT $m$]

Score: [SCORE $m$]

\vspace{5pt}

Summarize these explanation and generate a new prompt accordingly. Put the new prompt within tag $<$prompt$>$ and $<$/prompt$>$.

\end{tcolorbox}

\clearpage
\newpage

\subsection{Detailed Results on LCP Ablations}

Table~\ref{table:contrastive_ablation}
and Table~\ref{table:num_prompts_ablation} show detailed results on accuracies and win rates over the 17 tasks for the BBH data for the two ablation studies: w/ and w/o contrastive learning, and number of generated prompt candidates, respectively.

\begin{table}[h]
\small
\caption{Test accuracy of LCP with and w/o contrastive learning on four types of 17 BBH
tasks for last iteration prompt (Last) versus the prompt with best training accuracy (Best). \textcolor{blue}{Blue} indicates overall best win rates for Last or Best. }
\label{table:contrastive_ablation}
\begin{center}
\begin{tabular}{lcc}
\hline
\multicolumn{1}{c}{\bf TASK}  &\multicolumn{1}{c}{\bf LCP} &\multicolumn{1}{c}{\bf LCP (w/o contrastive learning)}  \\
\hline
& \multicolumn{1}{c}{Last / Best} & \multicolumn{1}{c}{Last / Best}  \\
\hline
\emph{Algorithmic and Multi-Step Arithmetic Reasoning} &  \\
\hline
geometric\_shapes & 51.25 / \textbf{61.00} & \textbf{53.50} / 56.00\\
logical\_deduction\_three\_objects & {92.50} /  {90.25}  & \textbf{93.00} / \textbf{93.00} \\
logical\_deduction\_five\_objects &  \textbf{71.50} /  \textbf{74.50}  & 71.00 / 70.00 \\
logical\_deduction\_seven\_objects &  \textbf{62.00} /  \textbf{62.50} & 58.00 / 53.00 \\
penguins\_in\_a\_table &  \textbf{97.86} /  \textbf{96.15} &  94.23 / 93.87 \\
reasoning\_about\_colored\_objects & \textbf{85.30} / \textbf{84.40} & 80.50 / 83.50 \\
temporal\_sequences &   \textbf{98.25} /  97.00 & 94.50 / \textbf{98.50} \\
tracking\_shuffled\_objects\_three\_objects & \textbf{92.00} / \textbf{99.00} & 89.50 / 90.00 \\
tracking\_shuffled\_objects\_five\_objects &  {90.50} /  \textbf{95.50} & \textbf{94.50} / 92.50 \\
tracking\_shuffled\_objects\_seven\_objects &  \textbf{92.10} /  \textbf{98.30} & 89.50 / 92.50 \\
\hline
Win Rate (\%) &  \textbf{70.00} / \textbf{80.00} & 30.00 / 20.00 \\
\hline
\emph{Natural Language Understanding} & \\
\hline
disambiguation\_qa &  {66.83} /  \textbf{66.33}  & \textbf{70.50} / 63.50 \\
hyperbaton &  {78.25} /  \textbf{84.00}  & \textbf{79.00} / 79.50 \\
salient\_translation\_error\_detection & 57.25 /  \textbf{69.50} & \textbf{66.50} / 66.00 \\
snarks & 65.73 / \textbf{70.98} & \textbf{67.73} / 68.23 \\
\hline
Win Rate (\%) & 0.00 / \textbf{100.00} & \textbf{100.00} / 0.00  \\
\hline
\emph{Use of World Knowledge} &  \\
\hline
date\_understanding & 75.50 / 74.50  &  \textbf{77.00} / \textbf{75.00} \\
movie\_recommendation &  \textbf{87.75} /  \textbf{85.50} & 85.00 / 75.00 \\
ruin\_names & 76.50 / 75.25 & \textbf{77.50} / \textbf{78.00} \\
\hline
Win Rate (\%) & 33.00 / 33.00 & \textbf{67.00} / \textbf{67.00} \\
\hline
\emph{Multilingual Knowledge and Reasoning} &  \\
\hline
salient\_translation\_error\_detection & 57.25 /  \textbf{69.50} & \textbf{66.50} / 66.00  \\
\hline
Win Rate (\%) & 0.00 /  \textbf{100.00} & \textbf{100.00} / 0.00 \\
\hline
\textbf{Overall Win Rate (\%)} & 47.06 / \textcolor{blue}{\textbf{76.47}} & \textcolor{blue}{\textbf{52.94}} / 23.53 \\ \hline
\end{tabular}
\end{center}
\end{table}
\begin{table}[t]
\small
\caption{Test accuracy of LCP with different values of number of prompt candidates ($N$) on four types of 17 BBH
tasks for last iteration prompt (Last) versus the prompt with best training accuracy (Best). \textcolor{blue}{Blue} indicates overall best win rates for Last or Best. }
\label{table:num_prompts_ablation}
\begin{center}
\begin{tabular}{lcccc}
\hline
\multicolumn{1}{c}{\bf TASK}  & \multicolumn{1}{c}{\bf $N=10$} & \multicolumn{1}{c}{\bf $N=6$} & \multicolumn{1}{c}{\bf $N=4$} & \multicolumn{1}{c}{\bf $N=2$}\\
\hline
& \multicolumn{1}{c}{Last / Best} & \multicolumn{1}{c}{Last / Best} & \multicolumn{1}{c}{Last / Best} &  \multicolumn{1}{c}{Last / Best} \\
\hline
\multicolumn{3}{l}{\emph{Algorithmic and Multi-Step Arithmetic Reasoning}} \\
\hline
geometric\_shapes & 51.25 / \textbf{61.00} & \textbf{54.00} / 58.50 & 50.50 / 60.00 & 52.00 / 56.50 \\
logical\_deduction\_three\_objects &  \textbf{92.50} /  {90.25}  & 84.00 / \textbf{90.50} & 87.00 / 87.00 & 82.00 / 80.50 \\
logical\_deduction\_five\_objects &  \textbf{71.50} /  \textbf{74.50}  & 62.00 / 62.00 & 59.00 / 59.00 & 60.00 / 60.50 \\
logical\_deduction\_seven\_objects &  {62.00} /  \textbf{62.50} & \textbf{63.00} / 58.50 & 59.00 / 55.00 & 54.50 / 55.00  \\
penguins\_in\_a\_table &  \textbf{97.86} /  {96.15} &  94.87 / 94.87  & 96.58 / \textbf{96.58} & 95.73 / 95.73 \\
reasoning\_about\_colored\_objects & 85.30 / 84.40 & 83.00 / \textbf{87.00} & \textbf{86.50} / 83.50 & 82.50 / 82.50 \\
temporal\_sequences &   {98.25} /  {97.00} & \textbf{99.50} / 96.50 & 96.00 / 96.00 & 95.50 / \textbf{99.50} \\
tracking\_shuffled\_objects\_three\_objects & 92.00 / \textbf{99.00} & 94.50 / 94.50 & 94.00 / \textbf{99.00} & \textbf{95.50} / 95.50 \\
tracking\_shuffled\_objects\_five\_objects &  {90.50} /  \textbf{95.50} & 78.50 / 78.50 & 92.50 / 92.50 & \textbf{98.50} / 95.00 \\
tracking\_shuffled\_objects\_seven\_objects &  {92.10} /  \textbf{98.30} & \textbf{93.50} / 96.50 & 85.00 / 88.50 & 79.00 / 79.00  \\
\hline
Win Rate (\%) &  \textbf{63.00} / \textbf{83.00} & 60.00 / 47.00 & 43.00 / 37.00 & 33.00 / 27.00 \\
\hline
\emph{Natural Language Understanding} & \\
\hline
disambiguation\_qa &  {66.83} /  {66.33}  & 66.00 / 68.00 & \textbf{68.00} / \textbf{71.00} & 66.00 / 63.00 \\
hyperbaton &  {78.25} /  \textbf{84.00}  & 82.00 / 81.50 & 82.00 / {82.50} & 81.50 / \textbf{83.50} \\
salient\_translation\_error\_detection & 57.25 /  {69.50} & \textbf{70.00} / \textbf{70.00} & 65.50 / 68.50 & 67.50 / 65.50 \\
snarks & 65.73 / 70.98 & 60.14 / 60.14 & \textbf{74.13} / \textbf{76.22} & 71.33 / 71.33 \\
\hline
Win Rate (\%) & 25.00 / 58.00  & 42.00 / 42.00 & \textbf{75.00} / \textbf{67.00} & 42.00 / 33.00 \\
\hline
\emph{Use of World Knowledge} &  \\
\hline
date\_understanding & \textbf{75.50} / 74.50  &  72.00 / 72.00 & 73.50 / 74.00 & 74.00 / \textbf{76.00} \\
movie\_recommendation &  \textbf{87.75} /  {85.50} & 86.50 / \textbf{87.50} & 79.00 / 77.00 & 82.00 / 83.00  \\
ruin\_names & 76.50 / 75.25 & 76.00 / 79.50 & 79.00 / \textbf{80.00} & \textbf{80.00} / 77.50\\
\hline
Win Rate (\%) & \textbf{78.00} / 44.00  & 22.00 / \textbf{56.00} & 33.00 / 44.00 & 67.00 / \textbf{56.00} \\
\hline
\emph{Multilingual Knowledge and Reasoning} &  \\
\hline
salient\_translation\_error\_detection & 57.25 /  {69.50} & \textbf{70.00} / \textbf{70.00} & 65.50 / 68.50 & 67.50 / 65.50 \\
\hline
Win Rate (\%) & 0.00 / 67.00 & \textbf{100.0} / \textbf{100.0} & 33.00 / 33.00 & 67.00 / 0.00  \\
\hline
 \textbf{Overall Win Rate (\%)} & \textcolor{blue}{\textbf{56.86}} / \textcolor{blue}{\textbf{70.59}}  & 49.02 / 47.06 & 49.02 / 45.10 & 41.18 / 33.33\\ \hline
\end{tabular}
\end{center}
\end{table}

\newpage

\subsection{Detailed Results of Model Adaptation}

\begin{table}[t]
\small
\caption{Comparison of prompt adaptation and prompt optimization on Claude-3-Sonnet from Claude-3-Haiku: accuracy on BBH tasks for last iteration prompt versus best prompt on the training set. \textcolor{blue}{Blue} indicates overall best win rates for Last or Best.}
\label{table:cross_version_perf_sonnet}
\begin{center}
\begin{tabular}{lccc}
\hline
\multicolumn{1}{c}{\bf TASK}  &\multicolumn{1}{c}{\bf LCP Adaptation} &\multicolumn{1}{c}{\bf LCP Optimization} &\multicolumn{1}{c}{\bf Haiku Optimized}\\
\hline
& \multicolumn{1}{c}{Last / Best} & \multicolumn{1}{c}{Last / Best} & \multicolumn{1}{c}{Last / Best}\\
\hline
\multicolumn{3}{l}{\emph{Algorithmic and Multi-Step Arithmetic Reasoning}} \\
\hline
geometric\_shapes & \textbf{70.50} / \textbf{68.00} & 44.80 / 63.40 & 44.50 / 50.50 \\
logical\_deduction\_three\_objects & \textbf{95.50} / \textbf{91.50} & 90.80 / 91.30 & 65.00 / 87.50 \\
logical\_deduction\_five\_objects & 61.00 / \textbf{71.50} & \textbf{70.50} / 70.00 & 51.00 / 62.00 \\
logical\_deduction\_seven\_objects & 59.50 / \textbf{58.00} & 62.80 / 63.60 & \textbf{82.50} / 57.50 \\
penguins\_in\_a\_table & 93.20 / \textbf{95.70} & \textbf{97.20} / 94.40 & 85.00 / \textbf{95.70} \\
reasoning\_about\_colored\_objects & 82.50 / 84.00 & \textbf{85.30} / 84.50 & 66.50 / \textbf{84.50} \\
temporal\_sequences & 96.50 / 97.00 & \textbf{96.80} / 95.50 & 25.50 / \textbf{98.00} \\
tracking\_shuffled\_objects\_three\_objects & 91.00 / 96.50 & \textbf{96.80} / \textbf{98.80} & 84.50 / 97.00 \\
tracking\_shuffled\_objects\_five\_objects & 93.00 / 93.50 & 94.90 / \textbf{95.70} & \textbf{95.70} / 94.50 \\
tracking\_shuffled\_objects\_seven\_objects & \textbf{94.50} / 96.50 & 92.10 / \textbf{98.30} & 79.50 / 89.00 \\
\hline
Win Rates (\%) & 55.00 / 55.00 & \textbf{75.00} / \textbf{65.00} & 20.00 / 40.00 \\
\hline
\emph{Natural Language Understanding} & & \\
\hline
disambiguation\_qa & 61.50 / \textbf{73.50} & 67.00 / 61.10 & \textbf{72.50} / 70.50 \\
hyperbaton & \textbf{82.00} / \textbf{83.50} & 70.00 / 59.30 & 69.00 / 76.50 \\
salient\_translation\_error\_detection & \textbf{68.50} / \textbf{69.00} & 53.10 / 42.40 & 63.60 / 65.00 \\
snarks & 66.40 / \textbf{76.20} & 47.90 / 51.00 & \textbf{99.50} / 66.40 \\
\hline
Win Rates (\%) & \textbf{62.50} / \textbf{100.0} & 25.00 / 0.00 & \textbf{62.50} / 50.00 \\
\hline
\emph{Use of World Knowledge} & & \\
\hline
date\_understanding & 73.50 / 72.50 & 75.50 / 56.50 & \textbf{97.50} / \textbf{73.00} \\
movie\_recommendation & 51.00 / \textbf{87.50} & 75.90 / 78.90 & \textbf{91.00} / 86.50 \\
ruin\_names & 76.50 / 79.50 & 65.90 / 69.30 & \textbf{87.00} / \textbf{80.00} \\
\hline
Win Rates (\%) & 16.67 / 66.67 & 33.33 / 0.00 & \textbf{100.0} / \textbf{83.33} \\
\hline
\emph{Multilingual Knowledge and Reasoning} & & \\
\hline
salient\_translation\_error\_detection & \textbf{68.50} / \textbf{69.00} & 53.10 / 42.40 & 63.60 / 65.00 \\
\hline
Win Rates (\%) & \textbf{100.0} / \textbf{100.0} & 0.00 / 0.00 & 50.00 / 50.00 \\
\hline
\textbf{Overall Win Rates (\%)} & 50.00 / \textcolor{blue}{\textbf{67.65}} & \textcolor{blue}{\textbf{55.88}} / 38.24 & 44.12 / 50.00 \\
\hline
\end{tabular}
\end{center}
\end{table}

\begin{table}[t]
\small
\caption{Comparison of prompt adaptation and prompt optimization on Claude-3-Haiku from Claude-3-Sonnet: accuracy on BBH tasks for last iteration prompt versus best prompt on the training set.}
\label{table:cross_version_perf_haiku}
\begin{center}
\begin{tabular}{lccc}
\hline
\multicolumn{1}{c}{\bf TASK}  &\multicolumn{1}{c}{\bf LCP Adaptation} &\multicolumn{1}{c}{\bf LCP Optimization} &\multicolumn{1}{c}{\bf Sonnet Optimized} \\
& \multicolumn{1}{c}{Last / Best} & \multicolumn{1}{c}{Last / Best} & \multicolumn{1}{c}{Last / Best}\\
\hline
\multicolumn{3}{l}{\emph{Algorithmic and Multi-Step Arithmetic Reasoning} } \\
\hline
geometric\_shapes & 51.50 / 51.60 & 46.00 / 52.50 & \textbf{71.50} / \textbf{54.00} \\
logical\_deduction\_three\_objects & \textbf{76.00} / \textbf{78.50} & 70.00 / 68.00 & 67.00 / 76.50 \\
logical\_deduction\_five\_objects & 50.00 / 52.50 & \textbf{55.00} / 50.00 & 47.50 / \textbf{54.50} \\
logical\_deduction\_seven\_objects & 42.50 / \textbf{47.00} & 7.00 / 41.00 & \textbf{87.00} / 43.50 \\
penguins\_in\_a\_table & \textbf{82.90} / \textbf{86.30} & 79.50 / 82.90 & 78.00 / 82.90 \\
reasoning\_about\_colored\_objects & 64.50 / 66.50 & \textbf{67.50} / \textbf{69.00} & 53.50 / 67.00 \\
temporal\_sequences & 83.50 / 91.80 & \textbf{93.00} / \textbf{94.20} & 44.50 / 88.50 \\
tracking\_shuffled\_objects\_three\_objects & 64.00 / 66.00 & \textbf{67.00} / 66.00 & 66.00 / \textbf{73.50} \\
tracking\_shuffled\_objects\_five\_objects & 43.00 / 71.00 & 67.50 / 64.00 & \textbf{85.50} / \textbf{71.50} \\
tracking\_shuffled\_objects\_seven\_objects & 55.00 / 60.00 & \textbf{62.50} / \textbf{64.00} & 58.00 / 62.50 \\
\hline
Win Rates (\%) & 40.00 / 50.00 & \textbf{70.00} / 45.00 & 40.00 / \textbf{65.00} \\
\hline
\emph{Natural Language Understanding} & & \\
\hline
disambiguation\_qa & 67.00 / \textbf{66.00} & 61.00 / 63.50 & \textbf{75.00} / 65.50 \\
hyperbaton & 87.00 / 86.50 & \textbf{88.00} / 86.50 & 52.50 / \textbf{88.00} \\
salient\_translation\_error\_detection & 56.50 / 54.00 & 51.50 / 53.50 & \textbf{67.80} / \textbf{54.50} \\
snarks & 68.50 / \textbf{72.70} & 59.40 / 69.90 & \textbf{87.00} / 71.30 \\
\hline
Win Rates (\%) & 50.00 / \textbf{75.00} & 25.00 / 12.50 & \textbf{75.00} / \textbf{75.00} \\
\hline
\emph{Use of World Knowledge} & & \\
\hline
date\_understanding & \textbf{79.50} / 66.00 & 24.00 / 69.00 & 68.00 / \textbf{74.00} \\
movie\_recommendation & 73.50 / 72.50 & \textbf{80.00} / \textbf{78.00} & 67.00 / 69.50 \\
ruin\_names & 48.50 / 65.00 & 60.50 / \textbf{72.50} & \textbf{63.50} / 65.00 \\
\hline
Win Rates (\%) & \textbf{50.00} / 33.33 & 50.00 / \textbf{83.33} & \textbf{50.00} / 50.00 \\
\hline
\multicolumn{3}{l}{\emph{Multilingual Knowledge and Reasoning}} \\
\hline
salient\_translation\_error\_detection & 56.50 / 54.00 & 51.50 / 53.50 & \textbf{67.80} / \textbf{54.50} \\
\hline
Win Rates (\%) & 50.00 / 50.00 & 0.00 / 0.00 & \textbf{100.0} / \textbf{100.0} \\
\hline
\textbf{Overall Win Rates (\%)} & 44.12 / 52.94 & \textcolor{blue}{\textbf{55.88}} / 44.12 & 50.00 / \textcolor{blue}{\textbf{64.71}}\\
\hline
\end{tabular}
\end{center}
\end{table}

\begin{table}[t]
\caption{Comparison of prompt adaptation and prompt optimization on from Claude 3 Sonnet $\rightarrow$ LLAMA 3. : accuracy on BBH tasks for last iteration prompt versus best prompt on the training set. \textcolor{blue}{Blue} indicates overall best win rates for Last or Best.}
\label{table:llama-adapatation}
\begin{center}
\begin{tabular}{lccc}
\hline
\multicolumn{1}{c}{\bf TASK}  &\multicolumn{1}{c}{\bf LCP adaptation} &\multicolumn{1}{c}{\bf LCP optimization} &\multicolumn{1}{c}{\bf Sonnet Optimized} \\
\hline
& \multicolumn{1}{c}{Last / Best} & \multicolumn{1}{c}{Last / Best} & \multicolumn{1}{c}{Last / Best}\\
\hline
\multicolumn{3}{l}{\emph{Algorithmic and Multi-Step Arithmetic Reasoning}} \\
\hline
geometric\_shapes & 21.00 / \textbf{30.50} & 10.50 / 21.50 & \textbf{68.50} / 27.50 \\
logical\_deduction\_three\_objects & \textbf{68.00} / \textbf{83.50} & 67.50 / 79.00 & 57.00 / 73.00 \\
logical\_deduction\_five\_objects & 38.50 / 53.00 & \textbf{53.00} / 50.00 & 28.50 / \textbf{56.50} \\
logical\_deduction\_seven\_objects & 40.50 / \textbf{50.50} & 37.00 / 43.00 & \textbf{58.00} / 42.50 \\
penguins\_in\_a\_table & 59.80 / 66.70 & 67.50 / 72.60 & \textbf{73.50} / \textbf{73.50} \\
reasoning\_about\_colored\_objects & 66.00 / 62.50 & \textbf{67.50} / 59.50 & 49.50 / \textbf{68.00} \\
temporal\_sequences & 78.00 / 85.50 & \textbf{78.50} / \textbf{88.00} & 45.00 / 86.50 \\
tracking\_shuffled\_objects\_three\_objects & 57.50 / 68.00 & \textbf{65.50} / 63.00 & 41.50 / \textbf{71.50} \\
tracking\_shuffled\_objects\_five\_objects & 72.00 / \textbf{77.00} & 76.50 / 76.50 & \textbf{78.60} / 69.00 \\
tracking\_shuffled\_objects\_seven\_objects & 38.00 / 53.50 & 59.00 / \textbf{59.50} & \textbf{71.50} / 58.00 \\
\hline
Win Rate (\%) & 40.00 / \textbf{55.00} & \textbf{60.00} / 40.00 & 50.00 / \textbf{55.00} \\
\hline
\multicolumn{3}{l}{\emph{Natural Language Understanding}}  & \\
\hline
disambiguation\_qa & 56.50 / 62.50 & 51.00 / \textbf{63.50} & \textbf{67.00} / 59.50 \\
hyperbaton & 52.50 / 55.50 & \textbf{53.50} / \textbf{60.50} & 36.00 / 60.00 \\
salient\_translation\_error\_detection & 40.00 / 44.00 & 37.50 / \textbf{48.00} & \textbf{55.20} / 25.50 \\
snarks & 36.40 / 46.90 & 45.50 / \textbf{64.30} & \textbf{80.00} / 52.40 \\
\hline
Win Rate (\%) & 37.50 / 25.00 & 37.50 / \textbf{100.0} & \textbf{75.00} / 25.00 \\
\hline
\emph{Use of World Knowledge} & & & \\
\hline
date\_understanding & 58.50 / \textbf{69.50} & 42.50 / 62.50 & \textbf{71.00} / 67.00 \\
movie\_recommendation & 44.00 / \textbf{63.00} & 60.00 / 62.50 & \textbf{75.00} / 42.50 \\
ruin\_names & \textbf{72.00} / 70.50 & 55.00 / \textbf{74.50} & 69.00 / 65.00 \\
\hline
Win Rate (\%) & 50.00 / \textbf{83.33} & 16.67 / 50.00 & \textbf{83.33} / 16.67 \\
\hline
\multicolumn{3}{l}{\emph{Multilingual Knowledge and Reasoning}} & \\
\hline
salient\_translation\_error\_detection & 40.00 / 44.00 & 37.50 / \textbf{48.00} & \textbf{55.20} / 25.50 \\
\hline
Win Rate (\%) & 50.00 / 50.00 & 0.00 / \textbf{100.00} & \textbf{100.00} / 0.00 \\
\hline
\textbf{Overall Win Rate (\%)} & 41.18 / 52.94 & 47.06 / \textcolor{blue}{\textbf{55.88}} & \textcolor{blue}{\textbf{61.76}} / 41.18 \\
\hline
\end{tabular}
\end{center}
\end{table}
\clearpage
\newpage

 \subsection{Examples of Generated Prompts}
 \begin{center}
\begin{longtable}{p{5.5cm}p{1cm}p{8cm}}
\caption{Comparison of prompts generated by our proposed LCP, AutoHint and OPRO.}
\label{table:generated_prompts}
\\\hline
\multicolumn{1}{c}{\bf TASK}  &\multicolumn{1}{c}{\bf Method} &\multicolumn{1}{c}{\bf Generated Prompts} \\
\hline
geometric\_shapes & LCP & To determine the expected output shape from an SVG path element, follow these steps:

1. Parse the path data string and identify the distinct line segments formed by the path commands and coordinates.
2. Count the number of line segments, as each segment represents a side of the resulting polygon.
3. Map the number of sides to the corresponding polygon name (e.g., 3 sides = triangle, 4 sides = quadrilateral, 5 sides = pentagon, 6 sides = hexagon, 7 sides = heptagon, and so on).
4. Note that the path data typically starts with a moveTo (M) command to set the initial point, followed by lineTo (L) commands that define the sides of the polygon by connecting the vertices.

By following this process, you can accurately identify the expected output shape based on the number of sides determined from the SVG path data. \\
& AutoHint & The summarized reason for the expected outputs is that the SVG path element with the given "d" attribute value defines a series of line segments that connect specific points, forming a closed geometric shape. The coordinates specified in the "d" attribute determine the vertices or corners of the shape, and the sequence of line segments connecting these points outlines the overall geometry. By analyzing the number of sides or angles formed by the connected line segments, one can identify the type of polygon or shape being drawn, such as a kite, trapezoid, or pentagon.\\
& OPRO & The SVG path element specifies a sequence of line segments and coordinates that connect to form a closed polygonal shape. By analyzing the path data, which defines the vertices and line segments, the number of sides in the resulting polygon can be determined. For example, if the path data forms a closed shape with 3 vertices and 3 line segments, the expected output would be a triangle. Similarly, shapes with 4, 5, 6, or more sides would correspond to quadrilaterals, pentagons, hexagons, and so on, respectively. The task is to examine the path data and identify the type of polygon created based on the number of line segments and vertices in the closed shape.
\\
\hline
logical\_deduction\_three\_objects & LCP & 1) Carefully analyzing the given comparative statements that describe relationships or ordering among three objects.

2) Using logical reasoning and deductive skills to establish the complete order or ranking of the objects based on the provided information.

3) Identifying the object that satisfies the specific condition or criteria mentioned in the question (e.g., second-oldest, second-most expensive) based on the deduced order or ranking.

4) The reasons should explain how the correct option is determined by logically interpreting the statements and accurately reflecting the deduced order through careful analysis and reasoning.
\\
& AutoHint & The summarized reasons are: In these types of questions, the key is to carefully analyze the given statements and logically deduce the order or ranking of the objects/entities based on the provided information. The statements establish relationships between the objects, such as "newer than," "less expensive than," or "finished before/after." By piecing together these relationships, one can infer the correct order or ranking, eliminating the incorrect options and identifying the valid answer choice.
\\
& OPRO & To solve these problems, the crucial step is meticulously analyzing the provided statements that describe the comparisons or relative qualities among the three objects. By logically interpreting these clues and deducing their implications, we can establish the complete ordering or ranking of the objects. Once this order is determined, the task is to identify the object that satisfies the specified condition, such as being the second-most expensive or second-newest, to arrive at the correct solution.
\\
\hline
logical\_deduction\_five\_objects & LCP & To reliably solve ordering and sequence problems that require deducing the correct arrangement based on logical constraints, follow these steps:

1. Carefully read and analyze each given statement or condition to identify constraints on the ordering, such as explicit mentions of absolute positions (e.g. "X is the first"), relative comparisons (e.g. "Y is more expensive than Z"), or ranges (e.g. "A is one of the three cheapest").

2. Translate each statement into a logical constraint on the ordering and use reasoning to deduce partial orderings or relationships between objects or entities based on these constraints.

3. Systematically combine these partial orderings and relative relationships by considering all possible arrangements and eliminating any that contradict the given information.

4. Construct the complete sequence or ranking that satisfies all constraints simultaneously.

5. Map this deduced ordering to the specific condition asked in the question (e.g. third from the left, second most expensive) to determine the expected output or correct option.

By methodically identifying constraints, applying logical reasoning to deduce orderings, and mapping the solution to the requested condition, you can reliably solve these types of ordering problems.
\\
& AutoHint & The summarized reasons are: The expected output is based on the explicit statements or logical implications provided in the given information, which establish the correct ordering or relationship among the entities involved.
\\
& OPRO & To solve ordering problems based on logical statements, follow this systematic approach:

1. Carefully read through all the given statements that describe relationships between objects or their positions relative to each other or the ends.

2. Identify any direct statements about the exact position of an object (e.g., "X is the third from the left").

3. Look for comparative statements that indicate the order between two objects (e.g., "Y is to the right of Z").

4. Use the identified information to reconstruct the overall order by positioning objects based on their relationships and constraints.

5. While constructing the order, make valid logical inferences from the given information to deduce the positions of objects not directly specified.

6. Check if the deduced order satisfies all the given statements consistently without violating any constraints.

7. The option corresponding to this fully reconstructed order that meets all the conditions is the correct answer.

By meticulously following this step-by-step approach and making careful logical deductions, you can reliably solve ordering problems based on statements describing relative positions.
\\
\hline
logical\_deduction\_seven\_objects & LCP & To solve logical reasoning problems involving the order or relative positions of objects, follow this systematic approach:

1) Thoroughly read and comprehend all the given statements or constraints describing the relationships, positions, or orders of the objects.

2) Identify any definitive placements or orderings of objects that are explicitly stated in the constraints.

3) Use logical deduction to derive additional constraints, implications, or relationships based on the given information and the definitive placements/orderings identified in step 2.

4) Methodically evaluate each option, eliminating those that violate any of the stated constraints or logically deduced implications from step 3.

5) The remaining option(s) that adheres to all the given constraints and deduced implications is the expected correct output.

The key is to diligently analyze the information, recognize logical relationships between objects, employ deductive reasoning based on the constraints, and arrive at a solution that is consistent with all the provided information.
\\
& AutoHint & Understood, I will provide a general summary of the reasons for the expected outputs without referring to any specific examples or entities mentioned in the data.
\\
& OPRO & To solve logic problems involving the ordering of a set of objects, the key steps are:
1. Carefully read and analyze all the given statements describing the relative positions or characteristics of the objects.
2. Identify the logical constraints and relationships imposed by each statement, such as "X is to the left of Y" or "X is taller than Y".
3. Use logical reasoning to deduce the implications of these constraints on the positions of the objects relative to each other.
4. Systematically combine the deduced information to reconstruct the complete order while ensuring it satisfies all the provided conditions.
5. Eliminate any options that violate the inferred order or the given constraints.
6. The option that correctly reflects the deduced order based on the given information is the solution.
\\
\hline
penguins\_in\_a\_table & LCP & To solve a problem involving tabular data, one must carefully analyze the information presented in the given table(s). Identify the specific column(s) or data points that are relevant to answering the question. Then, perform any necessary operations on that data, such as sorting, filtering, counting, or calculations, as per the requirements stated in the question. After logically processing the relevant data, you can determine the correct answer choice or expected output.
\\
& AutoHint & The summarized general reason for the expected outputs is that they are based on carefully analyzing the given information or data and applying logical reasoning to arrive at the correct answer. The expected outputs are determined by thoroughly understanding the context, identifying the relevant details, and making deductions or inferences based on the provided facts or conditions.
\\
& OPRO & To effectively solve questions involving tabular data, carefully analyze the structure and contents of the given table(s). Identify the column(s) containing information pertinent to the question asked. Based on the requirements stated in the question, you may need to perform operations such as sorting the relevant column(s) in ascending or descending order, filtering the data based on certain criteria, counting specific occurrences, or calculating derived values using the data. Logically process the tabular data by applying the necessary operations, and use the resulting information to arrive at the correct answer choice.
\\
\hline
reasoning\_about\_colored\_objects & LCP & To determine the expected output, study the provided set of items and their descriptions (color, shape, size, etc.). Take note of the particular attribute or condition specified in the question, such as "items of a certain color" or "items remaining after removing a specific type." Systematically go through each item, checking if it fulfills the stated condition. Count the total number of items that meet the criteria. The option that matches this final count represents the expected output.
 \\
& AutoHint & The summarized reasons for the expected outputs are:
The questions provide information about a set of items arranged in a specific order or position relative to each other. The expected output is determined by carefully analyzing the given details, such as the colors of the items, their arrangement, and the specific item or position referenced in the question. By logically interpreting the spatial relationships and attributes described in the input, one can deduce the correct answer choice that satisfies the conditions stated in the question.
\\
& OPRO & To solve the problem accurately, carefully read the given information to identify the set of items or objects described, along with their relevant attributes (such as color, type, etc.). Understand the specific condition or operation mentioned in the question (e.g., removing certain items, counting items with a particular attribute). Apply this condition or operation to the identified set of items, modifying or filtering the set as instructed. Then, logically analyze the resulting set of items to determine the option that correctly matches the final composition or count after applying the specified condition.
\\
\hline
temporal\_sequences & LCP & The expected output represents the sole remaining time window that is not accounted for in the person's daily schedule and activities as described. It is determined by meticulously considering all the provided information about the person's whereabouts and commitments throughout the day, as well as any relevant constraints like opening/closing hours of the location. By systematically eliminating all the other time slots occupied by the person's observed activities or locations, the correct answer emerges as the only unoccupied period when the person could have potentially visited the specified destination.
 \\
& AutoHint & I will provide a general summary of the reasons for the expected outputs, without referring to any specific examples or entities mentioned in the data.
\\
& OPRO & The solution involves carefully examining the timeline of events and activities provided in the problem. First, identify all the time slots where the person's whereabouts and activities are explicitly stated. Then, determine the remaining time window that is not covered by any of these known activities or constraints, such as the operating hours of the location mentioned. This unoccupied time period represents the only available opportunity for the person to have visited the specified destination (e.g., bakery, library, movie theater) before it closed for the day. By process of elimination, this remaining time slot becomes the most logical answer for when the person could have gone to the location in question.
\\
\hline
tracking\_shuffled\_objects\_three\_objects & LCP & To solve problems involving a sequence of swaps or exchanges between multiple people, it is crucial to carefully track the movement of each item or position through the entire series of swaps. Begin by noting the initial state, mapping which person or entity holds which item or position. Then, systematically follow each swap or exchange step-by-step, updating the holdings or positions after each swap according to the provided sequence. By meticulously tracking these changes through the entire set of swaps, you can determine the final state and identify the correct answer corresponding to the item or position held by the person or entity in question after the last swap involving them has occurred.
 \\
& AutoHint & Understood, I will provide a general summary of the reasons for the expected outputs without referring to any specific examples or entities mentioned in the data.
\\
& OPRO & In these types of questions, there are typically several individuals (say, Alex, Maya, and Sameer) who are initially assigned certain roles or possessions (e.g., playing a sport position, holding a particular object). The problem then describes a sequence of swaps or trades between pairs of these individuals, where they exchange their roles or possessions. To determine the final role or possession of a specific individual after all the swaps, it is crucial to carefully follow the entire sequence, meticulously updating each individual's state after every swap. By accurately tracing the swaps from the initial assignments to the end, you can arrive at the correct final state or possession for the given individual.
\\
\hline
tracking\_shuffled\_objects\_five\_objects & LCP & In these types of problems involving swaps or exchanges, the key to arriving at the correct solution is to diligently track the sequence of changes that occur. By methodically following each swap or exchange step-by-step and updating the current state of assignments or positions, one can trace the path of how entities (people, objects, etc.) move from their initial state to the final outcome. Maintaining an organized record of the swaps and their effects allows you to meticulously trace the progression until the end, enabling you to determine the final configuration accurately.
 \\
& AutoHint & Understood, I will provide a general summary of the reasons for the expected outputs without referring to any specific examples or entities mentioned in the data.
\\
& OPRO & In these types of problems involving swaps or trades, it is crucial to first understand the initial distribution of objects among a group of individuals. Then, methodically trace each swap or trade that occurs between pairs of individuals, carefully updating the ownership of objects after each step. By systematically following the given sequence of swaps from start to end, updating who holds which object after each transaction, the final state or ownership of objects can be accurately determined.
\\
\hline
tracking\_shuffled\_objects\_seven\_objects & LCP & These types of problems involve tracking the movement of objects or assignments between a set of individuals through a given sequence of swaps or trades. The key to finding the correct solution is to start from the initial state where objects or roles are assigned to individuals, and then meticulously follow each swap step-by-step, updating the associations between individuals and objects/roles as dictated by the swap instructions. By carefully applying the entire sequence of swaps to the initial state, keeping an accurate record of how objects or assignments are redistributed among the individuals, the final state can be determined, revealing the correct output option for the object or role associated with a particular individual after all swaps have occurred.
 \\
& AutoHint & I will provide a general summary of the reasons for the expected outputs, without referring to any specific examples or entities mentioned in the data.
\\
& OPRO & In these types of problems involving swaps or exchanges of items (e.g., gifts, positions, partners) between individuals, the key to finding the solution is to diligently track the changes that occur after each swap. Start by understanding the initial state or arrangement of items among the individuals. Then, systematically follow the given sequence of swaps, updating the state by moving the items between individuals as specified. After applying all the swaps in the correct order, you will arrive at the final state, allowing you to determine the ultimate position or ownership of the particular item or individual in question.
\\
\hline
disambiguation\_qa & LCP & To correctly identify the antecedent (the noun or noun phrase that a pronoun refers to) in a sentence, follow these steps:

1. Examine the context and meaning of the sentence to understand the logical relationships and actions described.

2. Identify the nouns or noun phrases that the pronoun could potentially refer to based on grammatical agreement (number, gender, etc.) and proximity to the pronoun.

3. Evaluate each potential antecedent by substituting it for the pronoun. The antecedent should maintain the logical flow and coherence of the sentence.

4. If only one potential antecedent satisfies the criteria in step 3, that noun or noun phrase is the unambiguous antecedent.

5. If multiple potential antecedents satisfy the criteria, and the context does not provide enough information to disambiguate, then the pronoun usage is considered ambiguous.
 \\
& AutoHint & The summarized reasons for determining the antecedent of a pronoun in a sentence are as follows:
The pronoun typically refers to the subject or object of the main clause that logically connects to the clause containing the pronoun. The context and logical flow of the sentence provide clues to identify the antecedent. If the pronoun can reasonably refer to multiple entities mentioned in the sentence, then the antecedent is considered ambiguous due to a lack of sufficient contextual information.
\\
& OPRO & To correctly identify the antecedent (the entity that a pronoun refers to) within a sentence, it is crucial to analyze the context and relationships described. The pronoun should logically reference the most plausible noun or noun phrase based on the meaning conveyed by the sentence. Pay close attention to the surrounding information and flow of ideas to determine which entity performs or is associated with the actions mentioned. If there are multiple potential antecedents and the context lacks sufficient details to disambiguate, then the pronoun usage is considered ambiguous, as the referent cannot be definitively pinpointed.
\\
\hline
hyperbaton & LCP & In the English language, when multiple adjectives are used to describe a noun, they must follow a specific order to construct grammatically correct sentences. This conventional order is: Opinion, Size, Age, Shape, Color, Origin, Material, Qualifier/Purpose, Noun. Deviating from this standardized sequence results in unnatural and potentially incorrect phrasing.
 \\
& AutoHint & The summarized reason is: There are established conventions or rules for the correct order of adjectives when multiple adjectives are used to modify a noun. The expected output follows these conventions, ensuring that the adjectives are arranged in the proper sequence based on their specific categories or types.
\\
& OPRO & In the English language, when multiple adjectives are used to describe a noun, they are expected to follow a specific order for clear and natural sentence construction. This established order places opinion adjectives first, followed by size, age, shape, color, origin, material, and purpose adjectives modifying the noun. Adhering to this conventional adjective order is crucial for coherence and proper comprehension of the description.
\\
\hline
salient\_translation\_error\_detection & LCP & The expected output category should capture the type of error or change introduced in the English translation compared to the original German text. Consider the following error categories:

- Named Entities: Incorrect translation of proper names, locations, or other entities.

- Numerical Values: Missing, added, or altered numbers, dates, measurements, or numerical expressions.

- Modifiers/Adjectives: Changes to descriptive words, adjectives, or modifiers that alter the attributes or qualities of a noun.

- Negation/Antonyms: Introduction of negation, or swapping comparatives with their opposites/antonyms, altering the intended meaning.

- Trivial Factual Errors: Inaccuracies or mistakes in factual information unrelated to the other categories.

- Dropped Content: Significant omission of phrases, clauses, or parts of the original text in the translation.

Identify which of these error categories best describes the change or discrepancy observed in the given translation compared to the source German text. \\
& AutoHint & The summarized reasons for the expected outputs in the given examples are:
The errors in the translations can be categorized into different types, such as Named Entities, Numerical Values, Modifiers or Adjectives, Negation or Antonyms, Facts, and Dropped Content. The expected outputs identify the specific type of error present in each translation. The reasons provided explain how the translation deviates from the original meaning or content, leading to the identified error type. This could involve missing or altering crucial information like names, numerical values, modifiers, introducing negations or antonyms, factual inaccuracies, or omitting significant clauses or content from the original text.
\\
& OPRO & 1) Clearly stating that the expected output focuses on identifying the type of error introduced in the translation compared to the original text.

2) Listing and explaining the different categories of error types, such as changes to named entities, numerical values, modifiers/adjectives, negations/antonyms, factual errors, or dropped content.

3) Emphasizing that the expected output should correctly categorize the specific type of error present in the translation.
\\
\hline
snarks & LCP & Sarcasm relies on creating an intentional contradiction between the literal words used and the underlying sentiment being conveyed. It leverages techniques like hyperbole, irony, and rhetorical questioning to juxtapose opposing elements that clearly contradict common sense or reality. By expressing an exaggerated or mocking version of the opposite perspective, sarcastic statements unmask their true critical or derisive meaning beneath the facade of the contradictory words themselves. This discrepancy between the stated words and intended meaning is the hallmark of sarcastic communication.
 \\
& AutoHint & The summarized general reason for the expected sarcastic outputs in the given examples is that sarcasm is expressed through statements that contradict or exaggerate the intended meaning in an ironic or critical way. Sarcastic statements often convey the opposite of their literal meaning, using exaggeration, irony, or contradiction to imply criticism, mockery, or a different intended meaning than the literal words suggest.
\\
& OPRO & Sarcastic statements rely on creating a deliberate contradiction or contrast between the literal meaning and the intended meaning conveyed through irony or mockery. They often employ techniques like exaggeration, rhetorical questions, and juxtaposing positive/negative sentiments to highlight this incongruity. The sarcasm arises from this clash between the stated words and the true critical intent behind them, suggesting the opposite of what is expressed literally.
\\
\hline
date\_understanding & LCP & 1) Emphasize carefully analyzing the provided information, such as the current or starting date, time intervals (days, months, years), and any context about leap years.

2) Outline the key steps of establishing the reference date, calculating the target date by properly applying the specified time periods forward or backward, and handling factors like the number of days in each month and year boundaries.

3) Highlight the importance of paying close attention to details and performing accurate calculations to arrive at the correct date in the specified format (MM/DD/YYYY).
 \\
& AutoHint & The summarized general reason for the expected outputs is that the questions provide specific details about a date or event, and the correct answer corresponds to the date or day that logically follows from those details, taking into account the calendar system and conventions for representing dates.
\\
& OPRO & To accurately determine a date based on given information, it is crucial to methodically follow these steps:

1. Identify the provided reference date or starting point from the details given. This could be a birth date, anniversary, or specific calendar date.

2. Determine the time period or duration to calculate from the reference date. This may be a number of days, weeks, months, or years to be added or subtracted.

3. Consider if the time period should be added to the reference date to get a future date, or subtracted to get a past date. Carefully account for this direction.

4. Perform the date calculation, properly applying the time period while taking into account factors like number of days in each month and adjusting for leap years when necessary.

5. Ensure the final calculated date is presented in the exact format requested (e.g. MM/DD/YYYY).

By diligently analyzing all provided information and implementing precise step-by-step calculations while adhering to calendar conventions, the correct date can be determined reliably.
\\
\hline
movie\_recommendation & LCP & - Highly popular and critically acclaimed
- Culturally impactful and became a phenomenon

- Achieved mainstream success and global recognition

- From a comparable time period or era as the reference movies

- Represents a significant work in the context of popular cinema with broad appeal
 \\
& AutoHint & The summarized reasons for the expected outputs are:
The expected output is chosen because it shares similar genres, tones, themes, and overall cinematic styles with the given examples. The selected movie aligns with the general mood, narrative elements, and target audience of the reference films, making it the most appropriate choice among the provided options. Factors like genre (drama, action, thriller, etc.), tone (serious, lighthearted, suspenseful, etc.), and thematic elements (overcoming adversity, romance, historical events, etc.) are considered to determine the most suitable option that resonates with the given examples in terms of overall cinematic experience.
\\
& OPRO & The expected output is a movie that aligns closely with the examples provided in terms of genre (e.g. action, drama, comedy), tone/mood (e.g. lighthearted, gritty, emotional), level of critical praise and cultural significance, as well as overall production values and widespread appeal. The reasoning involves identifying the commonalities between the listed movies in terms of factors like storytelling approach, themes explored, filmmaking techniques, and target audience, then selecting the option that best matches that collective profile in a way that would be considered a comparable cinematic experience for viewers familiar with the given examples.
\\
\hline
ruin\_names & LCP & The expected output involves humorous edits that playfully modify the original names or phrases through clever linguistic techniques. These may include substituting a word with one that contrasts humorously, splitting words and recombining the parts to create new meanings, or introducing elements from wildly different contexts to generate an amusing, incongruous juxtaposition with the original. The key is to introduce an element of wordplay, unexpected meaning, or absurdity that creates a comedic effect, while still maintaining enough familiarity with the source material for the reader to recognize and appreciate the creative twist. \\
& AutoHint & The summarized reasons are: The expected outputs are considered humorous edits because they involve wordplay or puns created by slightly modifying the original word, phrase, or name in a clever or unexpected way. This can include replacing letters with similar-sounding ones, altering the spelling, or making slight changes to the wording. These types of edits are often used for comedic effect, as they play with the audience's familiarity with the original text while introducing a new, humorous interpretation or meaning.
\\
& OPRO & The expected outputs demonstrate clever and humorous modifications of familiar names, titles, or phrases. These edits playfully replace or alter certain words or letters to create an amusing contrast or incongruity with the original source material. Through techniques like wordplay, puns, and subtle linguistic substitutions, the humorous outputs inject an element of witty absurdity while still retaining a recognizable connection to the original. This form of intelligent and creative linguistic manipulation is an effective way to subvert expectations and elicit laughter by twisting the familiar into something comically unexpected.\\
\hline
\end{longtable}
\end{center}




\end{document}